\documentclass[lettersize,journal]{IEEEtran}
\usepackage{amsmath,amsfonts}
\usepackage{algorithmic}
\usepackage{algorithm}
\usepackage{array}
\usepackage[caption=false,font=normalsize,labelfont=sf,textfont=sf]{subfig}
\usepackage{textcomp}
\usepackage{stfloats}
\usepackage{url}
\usepackage{verbatim}
\usepackage{graphicx}
\usepackage{cite}
\usepackage{scalerel}
\usepackage{graphicx}
\usepackage{tikz}
\usepackage{hyperref}
\definecolor{orcidlogocol}{HTML}{A6CE39}

\newcommand{\orcidicon}{%
  \raisebox{0.7ex}{\scalerel*{\includegraphics[width=0.85em]{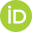}}{A}}%
}

\newcommand{\orcid}[1]{%
  \href{https://orcid.org/#1}{\textcolor{orcidlogocol}{\orcidicon}}%
}

\begin{document}

\title{ESACT: An End-to-End Sparse Accelerator for Compute-Intensive Transformers via Local Similarity}

\author{Hongxiang Liu\orcid{0009-0001-8278-3023}, 
        Zhifang Deng\orcid{0009-0005-3299-1616},
        Tong Pu\orcid{0009-0003-4068-649X},
        and Shengli Lu\orcid{0000-0001-5769-8671}
\thanks{The authors are with the School of Integrated Circuits, Southeast University, Nanjing 211189, China
(e-mail: liuhx@seu.edu.cn; zhifangdeng@seu.edu.cn; putong@seu.edu.cn; lsl@seu.edu.cn).
Corresponding author: Shengli Lu.}
}

\markboth{Journal of \LaTeX\ Class Files,~Vol.~14, No.~8, August~2021}%
{Shell \MakeLowercase{\textit{et al.}}: A Sample Article Using IEEEtran.cls for IEEE Journals}

\IEEEpubid{0000--0000/00\$00.00~\copyright~2021 IEEE}

\maketitle

\begin{abstract}
  Transformers, composed of QKV generation, attention computation, and FFNs, 
  have become the dominant model across various domains due to their outstanding performance. 
  However, their high computational cost hinders efficient hardware deployment. 
  Sparsity offers a promising solution, 
  yet most existing accelerators exploit only intra-row sparsity in attention, 
  while few consider inter-row sparsity. 
  Approaches leveraging inter-row sparsity often rely on costly global similarity estimation, 
  which diminishes the acceleration benefits of sparsity, 
  and typically apply sparsity to only one or two transformer components.
  Through careful analysis of the attention distribution and computation flow, 
  we observe that local similarity allows end-to-end sparse acceleration with lower computational overhead. 
  Motivated by this observation, we propose ESACT, 
  an end-to-end sparse accelerator for compute-intensive Transformers. 
  ESACT centers on the Sparsity Prediction with Local Similarity (SPLS) mechanism, 
  which leverages HLog quantization to accurately predict local attention sparsity prior to QK generation, 
  achieving efficient sparsity across all transformer components.
  To support efficient hardware realization, we introduce three architectural innovations. 
  First, we design a bit-level prediction unit that leverages bit-wise correlations 
  to enable efficient quantization and performs attention prediction using only additions, 
  significantly reducing power consumption. 
  Second, we propose a progressive generation scheme that overlaps QKV generation with sparsity prediction, 
  minimizing PE idle time. 
  Third, a dynamic allocation strategy is adopted to 
  alleviate computation imbalance caused by similarity-driven sparsity, improving overall PE utilization.
  Experimental results on 26 benchmarks demonstrate that 
  SPLS reduces total computation by 51.7\% with less than 1\% accuracy loss. 
  ESACT achieves an end-to-end energy efficiency of 3.27\,TOPS/W, 
  and improves attention-level energy efficiency by 2.95$\times$ and 2.26$\times$ over 
  SOTA attention accelerators SpAtten and Sanger, respectively.
\end{abstract}

\begin{IEEEkeywords}
Transformer, sparsity, local similarity, end-to-end, hardware accelerator.
\end{IEEEkeywords}

\section{Introduction}
\begin{figure}[t]
  \centering
  \includegraphics[width=\linewidth]{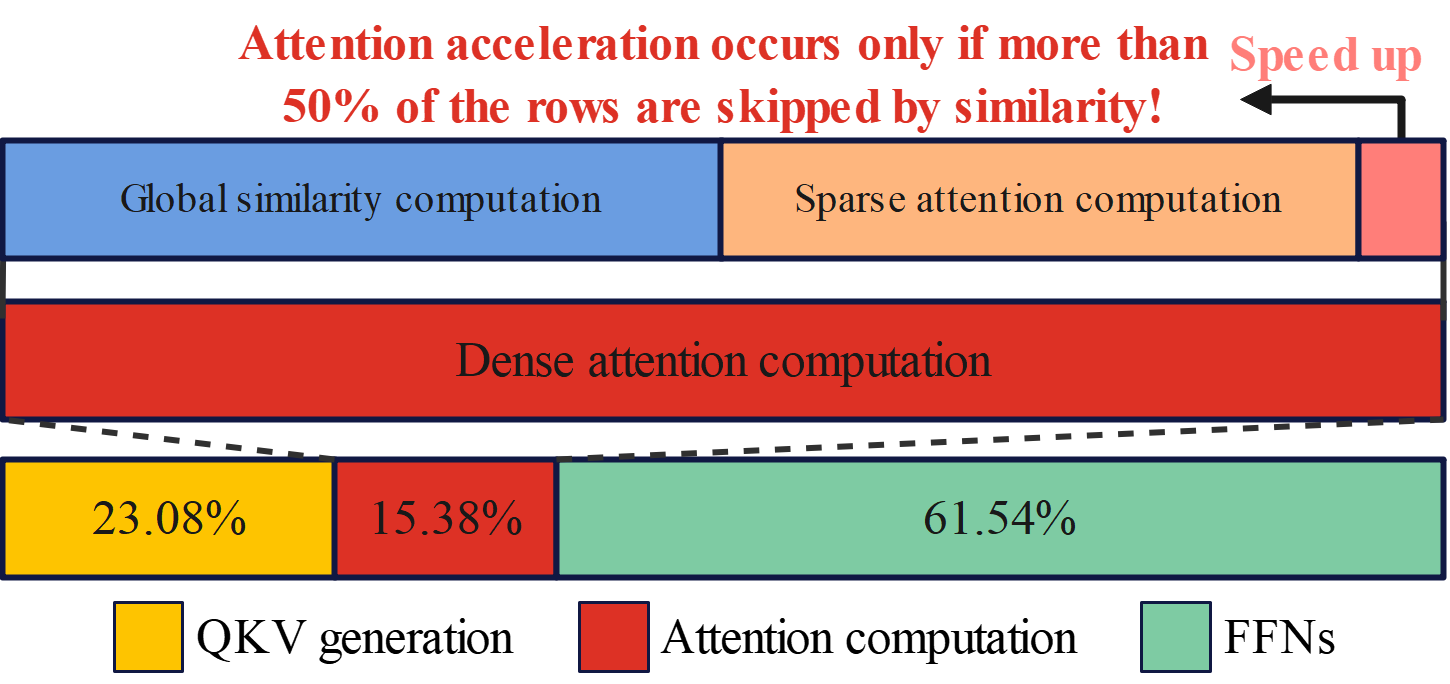}
  \caption{Computation breakdown of BERT-Large and the challenge of using global similarity for attention acceleration.}
  \label{fig:global}
\end{figure}

\IEEEPARstart{T}{ransformer}~\cite{trans} is a powerful neural network that has become a dominant model in various fields, 
including natural language processing~\cite{bert,gpt,t5,roberta,bart,mbart,ALBERT}, 
computer vision~\cite{vit,detr,swin,mae,seg,deit}, and time series forecasting~\cite{informer,autoformer,fedformer,crossformer,time}. 
Its remarkable performance is attributed to its innovative design features. 
Specifically, Transformer consists of two core components: the multi-head attention mechanism (MHA) and the feed-forward network (FFN). 
The MHA involves query, key and value (QKV) generation and attention computation, 
allowing the model to capture global dependencies while adaptively adjusting the weights based on the input. 
Subsequently, the FFN enhances the outputs from the MHA by performing feature extraction and transformation, 
which improves the model's capacity to abstract and represent information effectively. 
However, the performance improvement comes at the cost of increased computational complexity. 
As illustrated in Figure~\ref{fig:global}, taking BERT-Large~\cite{bert} as an example, when the input sequence length is 512, 
the total computation of the model is 167.5GFLOPs without any sparsity, 
of which the MHA and the FFN account for 38.46\% and 61.54\%, respectively. 
Therefore, it is essential to reduce the computation of the two core components simultaneously.

Sparsity is one of the most direct and effective approaches to addressing the computational bottleneck of a model 
by eliminating redundant operations, thereby reducing computational load. 
However, since sparsity removes certain computations, 
it may change the model's structure and potentially degrade its performance. 
Therefore, it is essential to apply sparsity in a way that minimizes computational cost while maintaining model accuracy.

We categorize existing effective sparsity methods for Transformers into two types: one based on the relative magnitude of data, 
and the other based on similarity relationships among data. 
The former is the most common sparsity approach~\cite{tf,spatten,fact,a3,sanger,elsa}, 
which typically predicts the attention matrix using quantized Q and K to obtain the predicted attention matrix (PAM). 
Importance is then evaluated based on the relative magnitudes of elements in the PAM, 
and a mask is generated accordingly to apply sparsity in the MHA or FFN during the actual computation stage. 
However, this method only considers intra-region relationships 
(e.g., selecting important positions by performing Top-$k$ within the row vectors of PAM) while neglecting inter-region relationships, 
thus limiting the sparsity potential of the model.

\IEEEpubidadjcol
In contrast, similarity-based sparsity~\cite{tsacc,sparc,interarch} leverages vector relationships 
by computing global similarity among row vectors of the attention matrix to identify redundant computations 
within the model. 
However, the computational cost of global similarity scales quadratically with the input sequence length, 
which severely impacts system latency. 
In some cases, the overhead introduced by computing global similarity may even exceed the computational savings gained through sparsity, 
leading to negative net benefits.

As shown in Figure~\ref{fig:global}, assuming an input sequence length of $l$ and neglecting intra-row sparsity from Top-$k$ pruning, 
if sparsification relies solely on global inter-row similarity in the attention matrix 
(and the cost of computing the similarity between two rows is assumed to be equivalent to computing a single attention score), 
then more than half of the rows (i.e., over $l/2$) must be sparsified to obtain any net performance gain. 
This requirement imposes a strict constraint on the effectiveness and scalability of similarity-based sparsity methods.
Moreover, existing accelerators employing similarity-based sparsity fail to sparsify all three components of the Transformer, 
thereby limiting their end-to-end acceleration performance.

Unlike previous work, we leverage local similarity to reduce the overhead of similarity prediction 
while enabling similarity-based sparsity across all three core components of the Transformer.
Our key insight is centered around the local attention matrix, which typically exhibits strong local similarity. 
This phenomenon arises from the observation that neighboring tokens often carry similar semantics. As a result, 
removing semantically redundant tokens within a local region does not significantly 
affect the overall semantic representation of that region.
However, existing quantization-based prediction methods either incur high computational and power overhead or 
fail to adequately preserve the original inter-row similarity of the attention matrix during prediction.

To address this issue and leverage the property of local similarity,
we propose a HLog-based Sparsity Prediction with Local Similarity (SPLS) mechanism. 
Specifically, we propose an efficient quantization method, HybridLog Quantization (HLog), 
which predicts the relative magnitudes of attention matrix elements 
while maximally preserving the original inter-row similarity prior to QK generation.
Subsequently, a row-wise top-$k$ pruning is applied to the PAM, resulting in the sparsified predicted attention (SPA). 
This step preserves the most important features in each row, not only eliminating redundant computations in attention, 
but also reducing the computational load in the subsequent local similarity matching stage.
Finally, local similarity analysis is performed on the SPA to identify critical vectors and their corresponding similar vectors. 
Based on the relationships between them, we perform structured sparsification 
across the three major computation modules of the Transformer.

Although the proposed SPLS mechanism can achieve significant end-to-end acceleration, 
its hardware implementation faces three main challenges.
First, the traditional method of quantizing data by 
individually comparing it with quantization levels limits the applicability of HLog quantization, 
resulting in considerable hardware overhead.
Second, to enable sparsity in QKV generation, 
a series of prediction operations must be performed prior to QKV generation, thereby increasing system latency.
Finally, since the similarity patterns vary across different heads, 
the concatenation of multi-head attention leads to uneven element distributions, 
which in turn reduces the utilization of PEs during runtime.

To address the above challenges, we propose the ESACT accelerator.
To the best of our knowledge, 
ESACT is the first accelerator that exploits local similarity to realize end-to-end sparsity across all transformer components.
The main contributions of this work are summarized as follows.

\begin{itemize}
\item We propose a HLog-based Sparsity Prediction with Local Similarity (SPLS) mechanism 
that effectively reduces the computational overhead of similarity prediction.
By leveraging HLog quantization for local attention prediction, 
it achieves end-to-end sparse acceleration while preserving the original similarity structure of the attention matrix.
\item We design a bit-level prediction unit that supports efficient hardware implementation. 
By exploiting bit-wise correlations for quantization and replacing multiplications with additions, 
it significantly reduces power consumption during attention prediction.
\item We propose a progressive generation scheme to reduce hardware system latency. 
By executing similarity prediction and QKV generation in parallel, it improves overall throughput. 
\item We develop a dynamic allocation strategy to address the irregular sparsity induced by similarity-based computation. 
By optimizing the critical path and similarity restoration process, 
it enhances PE array utilization and load balance.
\end{itemize}

Experimental results on 26 benchmarks demonstrate that 
SPLS reduces total computation by 51.7\% with less than 1\% accuracy loss. 
ESACT achieves an end-to-end energy efficiency of 3.27\,TOPS/W, 
and improves attention-level energy efficiency by 2.95$\times$ and 2.26$\times$ over 
SOTA attention accelerators SpAtten and Sanger, respectively.

The remainder of this paper is organized as follows.
Section~\ref{sec:background} introduces the basic structure of the Transformer and presents our design motivation.
Section~\ref{sec:software} provides a detailed description of the proposed SPLS mechanism, which enables end-to-end similarity-aware sparsity acceleration for Transformers.
Section~\ref{sec:hardware} elaborates on the architecture of the ESACT accelerator and presents three hardware innovations that ensure efficient hardware deployment.
Experimental results and conclusions are given in Sections~\ref{sec:eval} and ~\ref{sec:conclusion}, respectively.

\section{BACKGROUND AND MOTIVATION}\label{sec:background}

\subsection{Computation Flow of Transformers}

\begin{figure}[t]
  \centering
  \includegraphics[width=\linewidth]{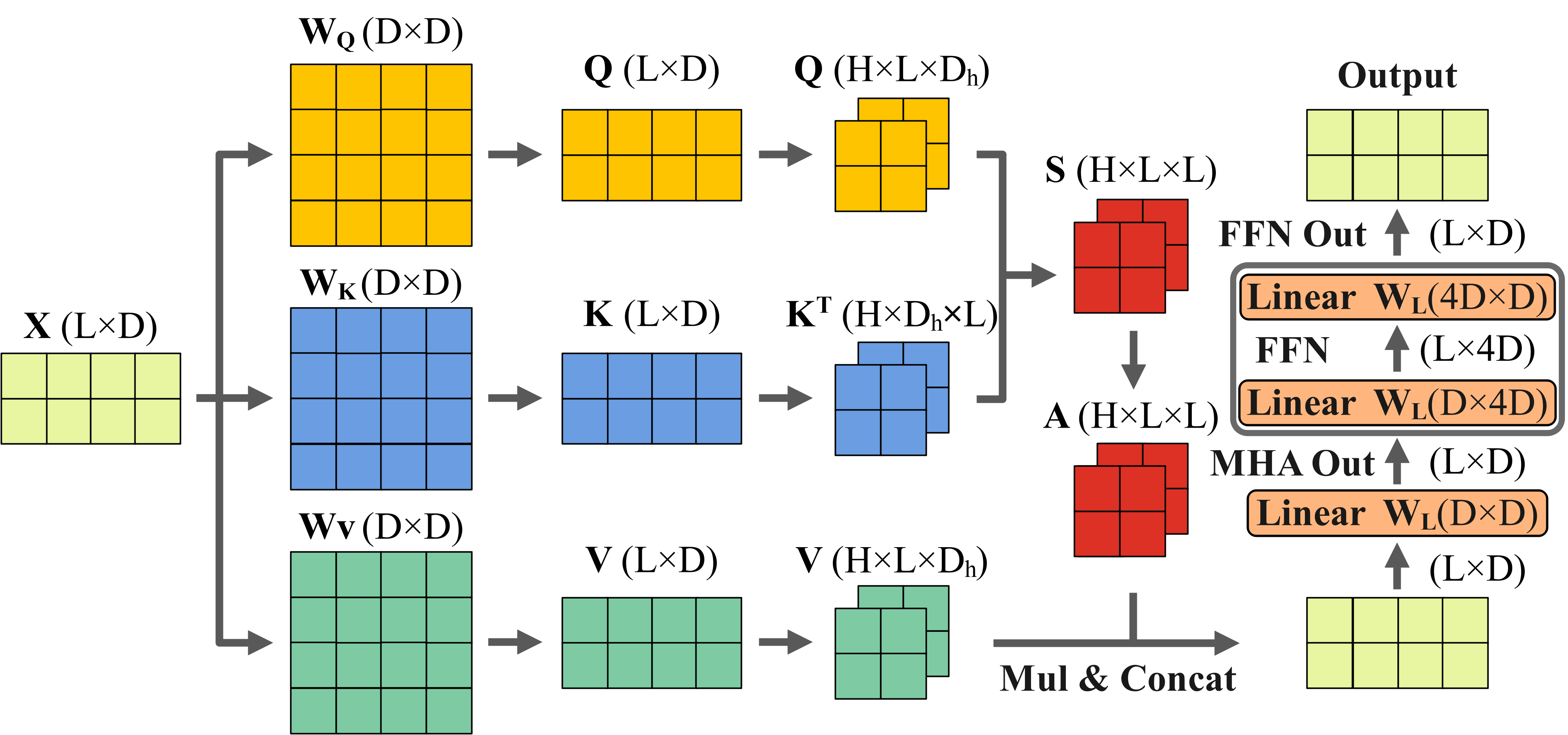}
  \caption{Computational flow of a Transformer block.}
  \label{fig:transformer_flow}
\end{figure}

The detailed computation flow of a Transformer block is illustrated in Figure~\ref{fig:transformer_flow}.
Given an input tensor of shape \( L \times D \), where L denotes the sequence length and D is the embedding dimension, 
three separate linear projections are first applied to generate the query (Q), key (K), 
and value (V), each with the same shape \( L \times D \). 
To support the MHA, Q, K, and V are each divided into H heads, 
resulting in individual head dimensions of \( L \times Dh \), where $D_h = D / H$.
Within each head, the attention computation is performed by multiplying Q with K$^{T}$ (the transpose of K), 
producing a score matrix S of shape \( L \times L \). 
S is then normalized row-wise using the softmax function to obtain the attention matrix A.
Next, A is multiplied by V, yielding the output of each head. 
The outputs from all heads are then concatenated into a single \( L \times D \) tensor and 
passed through a linear projection to restore the original dimensionality, resulting in the final MHA output.
The MHA output is subsequently fed into the FFN consisting of two linear transformations to 
produce the final output of the Transformer block. 
By stacking multiple such blocks, the Transformer is capable of 
capturing complex semantic relationships and enhancing its expressive capacity.

\begin{figure}[t]
  \centering
  \includegraphics[width=\linewidth]{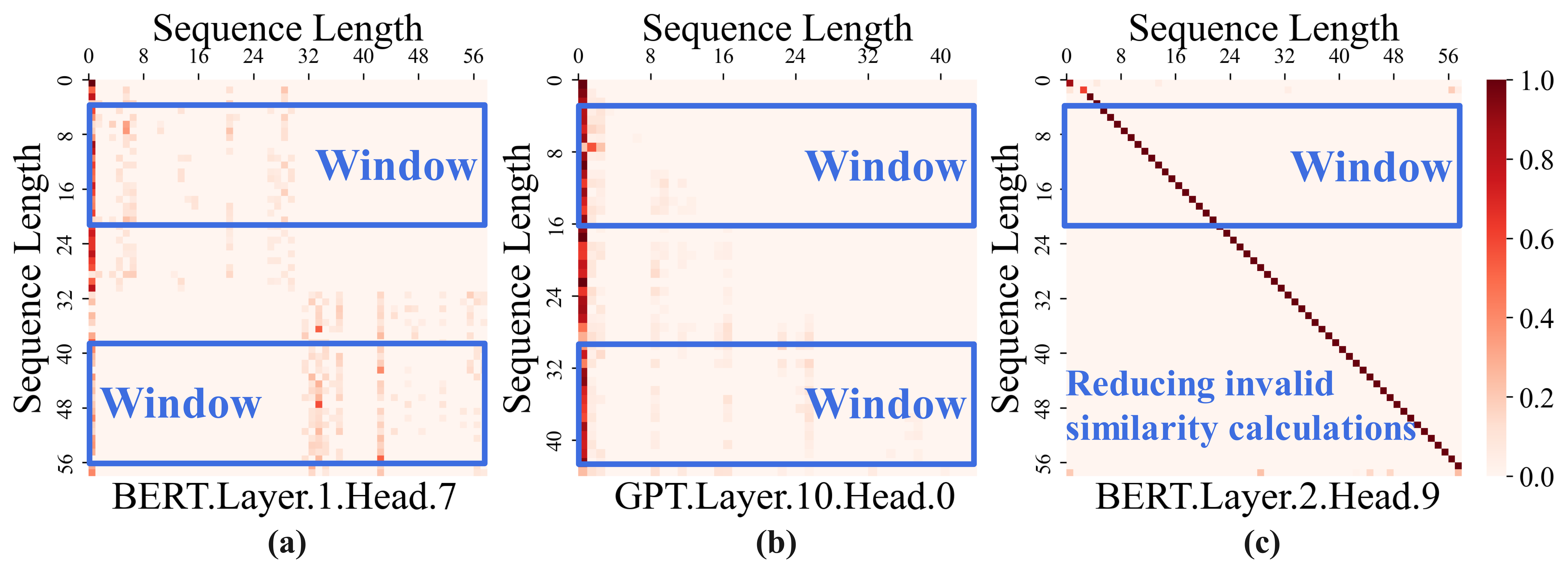}
  \caption{Visualization of the attention distribution.}
  \label{fig:kehsihua}
\end{figure}

\begin{figure}[t]
  \centering
  \includegraphics[width=\linewidth]{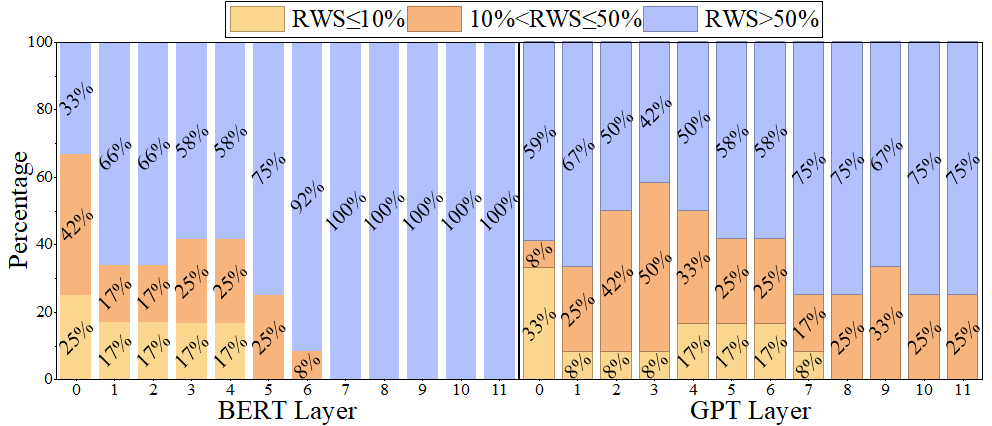}
  \caption{Percentage of heads exhibiting local similarity across different layers in BERT and GPT.}
  \label{fig:layerhead}
\end{figure}

\subsection{Motivation}

\begin{table*}[!htbp]
  \centering
  \renewcommand{\arraystretch}{1.5} 
  \caption{Comparison with existing sparse transformer accelerators.}
  \label{tab:sparse_acceleration_methods}
  \begin{tabular}{|c|c|c|c|c|c|c|c|}
    \hline
    \textbf{Accelerators} & 
    Sanger~\cite{sanger} & 
    SpAtten~\cite{spatten} & 
    DOTA~\cite{dota} &
    FACT~\cite{fact} & 
    TSAcc~\cite{tsacc} & 
    SpARC~\cite{sparc} &
    Our ESACT \\
    \hline
    \textbf{Sparse methods} & \multicolumn{4}{c|}{\textbf{Relative magnitude}} & 
    \multicolumn{2}{c|}{\textbf{Global similarity}} & 
    \textbf{Local similarity}\\
    \hline
    \textbf{Attn prediction methods} & 
    4-bit quant &
    Progressive quant &
    Low-rank &
    PoT quant &
    None &
    Low-rank &
    HLog quant\\
    \hline
    \textbf{Sparsity prediction cost} & 
    High &
    High &
    High &
    Low &
    High &
    High &
    Low\\
    \hline
    \textbf{Similarity fidelity} & 
    High &
    High &
    High &
    Low &
    None &
    High &
    High\\
    \hline
    \textbf{Sparse positions} & 
    Attn & 
    {\centering Attn \& FFN} &
    Attn &
    {\centering QKV \& Attn} & 
    QKV &
    Attn &
    {\centering QKV \& Attn \& FFN} \\
    \hline
  \end{tabular}
\end{table*}

\begin{figure*}[t]
  \centering
  \includegraphics[width=\linewidth]{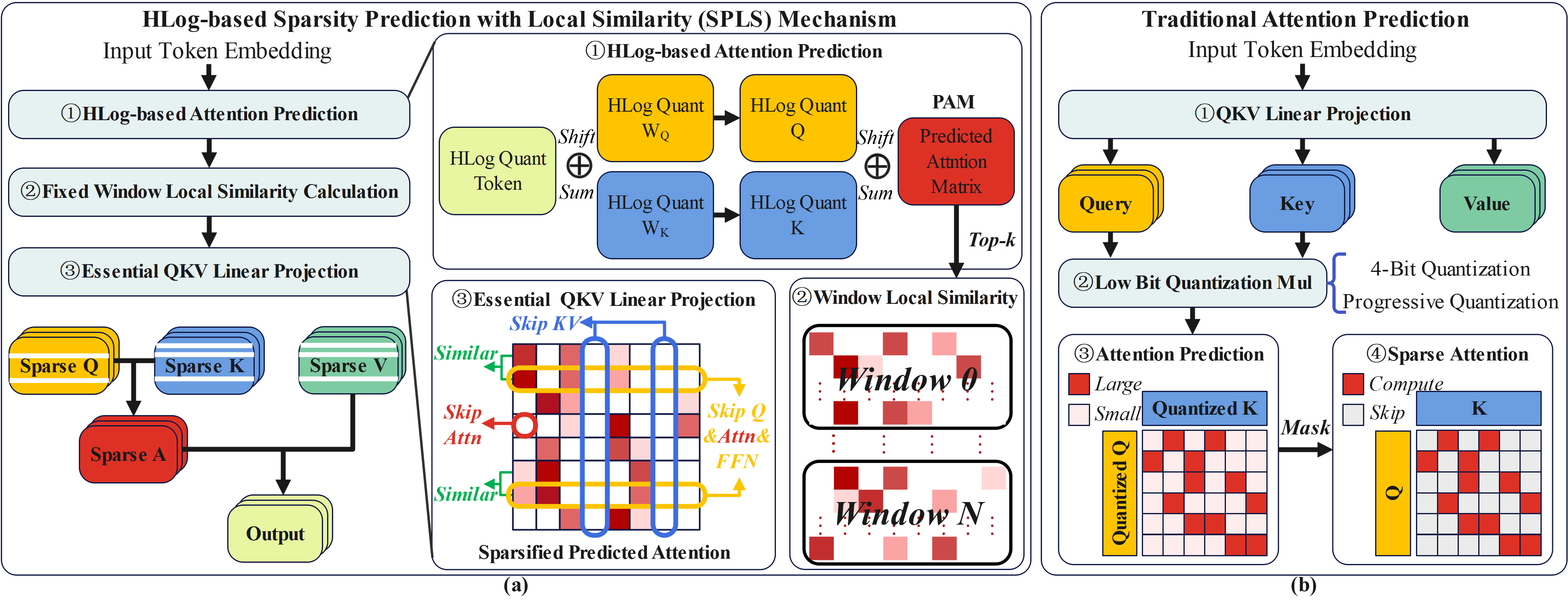}
  \caption{(a) Proposed SPLS mechanism. (b) Traditional attention prediction.}
  \label{fig:software}
\end{figure*}

Due to the inherent characteristics of natural language, semantically similar tokens are often located in nearby embedding spaces. 
For example, in the sentence ``Transformers adopt self-attention mechanisms to improve NLP tasks.'' 
the phrase ``self-attention mechanisms''  is semantically similar to ``Transformers''. 
As a result, redundant similarity can be removed without significantly altering the overall meaning of the sentence, 
yielding a simplified version like ``Transformers improve NLP tasks.''

To validate the effectiveness of local similarity, we apply sparsification to BERT and GPT-2 models 
on the MRPC~\cite{glue} and WikiText-2~\cite{wiki} datasets, 
respectively, while maintaining model accuracy. 
The resulting attention matrix distributions are illustrated in Figure~\ref{fig:kehsihua}.
It can be observed that after sparsification, 
the retained value positions of rows within the same local window, 
such as those in Figure~\ref{fig:kehsihua}(a) and (b), are almost identical, 
satisfying the prerequisite of inter-row similarity---similar data distributions across different rows.

To examine the prevalence of local similarity within the model, 
we measure the percentage of heads exhibiting local similarity across different layers in two models, 
as shown in Figure~\ref{fig:layerhead}.
Each head is divided into non-overlapping windows with a width of 8, 
and the heads are then grouped into three categories based on the ratio of windows exhibiting inter-row similarity (RWS).
It can be observed that most heads demonstrate a clear tendency of local inter-row similarity, 
strongly supporting our design motivation of leveraging local similarity.
Moreover, the prominent diagonal patterns in Figure~\ref{fig:kehsihua}(c) suggest that 
similarity computations are unnecessary in these heads.
By transforming global similarity computation into local similarity computation, 
the computational complexity is reduced from $l(l-1)/2$ to $l(w-1)/2$,
where $l$ is the sequence length and $w$ is the local window size. 
As a result, the adoption of local similarity effectively reduces unnecessary similarity computations and 
enables faster skipping of heads without inter-row similarity while 
adapting to the actual attention distribution.

Unfortunately, existing sparse Transformer accelerators 
fail to effectively exploit the inherent local similarity in attention to achieve end-to-end acceleration.
Table~\ref{tab:sparse_acceleration_methods} summarizes the sparsity methods adopted by several state-of-the-art accelerators.
Since attention is input-dependent, a prediction method is required to approximate the attention matrix, 
after which sparsification is performed either by comparing intra-row relative magnitudes or by evaluating inter-row similarity.
Sanger and SpAtten rely on low-bit quantization, and DOTA employs low-rank approximation for attention prediction. 
All of these methods require a large number of multiplications, resulting in substantial prediction overhead.
To reduce this cost, FACT adopts Power-of-Two (PoT) quantization to convert the prediction process into additions.
While effective at approximating relative element magnitudes, 
FACT struggles to preserve inter-row similarity and thus provides limited fidelity for similarity-based sparsity.
Moreover, although FACT applies mixed-precision computation in the FFN, 
it does not fully eliminate operations on irrelevant tokens, and therefore cannot realize true sparsity.
Methods that explicitly exploit inter-row sparsity, such as TSAcc and SpARC, 
compute global similarity across rows of the attention matrix.
This significantly increases the similarity-prediction overhead and limits system-level benefits.

Compared with these accelerators, 
our ESACT leverages HybridLog Quantization (HLog) to achieve high similarity fidelity with low prediction overhead.
By further utilizing local similarity to reduce the cost of similarity estimation, 
ESACT enables efficient, end-to-end sparse acceleration across the entire Transformer pipeline.

\section{ALGORITHM OPTIMIZATIONS}\label{sec:software}

Figure~\ref{fig:software}(a) illustrates the workflow of the HLog-based sparsity prediction with local similarity mechanism (SPLS), 
which consists of the following three steps.
First, we propose HybridLog Quantization (HLog) to perform attention prediction and obtain the PAM. 
Subsequently, a fixed-window similarity strategy is employed to perform local similarity computation 
on the sparsified predicted attention (SPA) derived from applying Top-$k$ pruning to the PAM. 
Finally, the similarity and sparsity characteristics of the SPA are utilized to guide the sparsification of the QKV generation, 
attention computation, and the FFN in the formal computation phase.
To better support hardware implementation, 
we further quantize all weights in the Transformer's linear transformations to 8-bit without sacrificing accuracy. 
All subsequent work is conducted based on this quantized model.

\subsection{HybridLog Quantization}\label{sec:qua}

Figure~\ref{fig:software}(b) presents the traditional attention prediction approach, 
in which the QK matrix is fully generated and subsequently quantized to facilitate attention prediction. 
This method, however, incurs substantial computational overhead due to the need to compute the entire QK matrix. 
In contrast, ESACT performs early prediction of essential QKV components prior to QK computation, 
thus significantly reducing unnecessary computation.
Quantization projects data values onto the nearest quantization levels.
Traditional 4-bit quantization~\cite{sanger} and progressive quantization~\cite{spatten} require multiplication operations, 
resulting in high computational cost. 
FACT~\cite{fact} and Enhance~\cite{enhance}, on the other hand, 
use PoT and APoT quantization to replace multiplication with more efficient shift-and-add operations.
However, these approaches suffer from either significant accuracy loss or overly complex control logic, 
which hinders their performance during the prediction phase.

\begin{figure}[t]
  \centering
  \includegraphics[width=\linewidth]{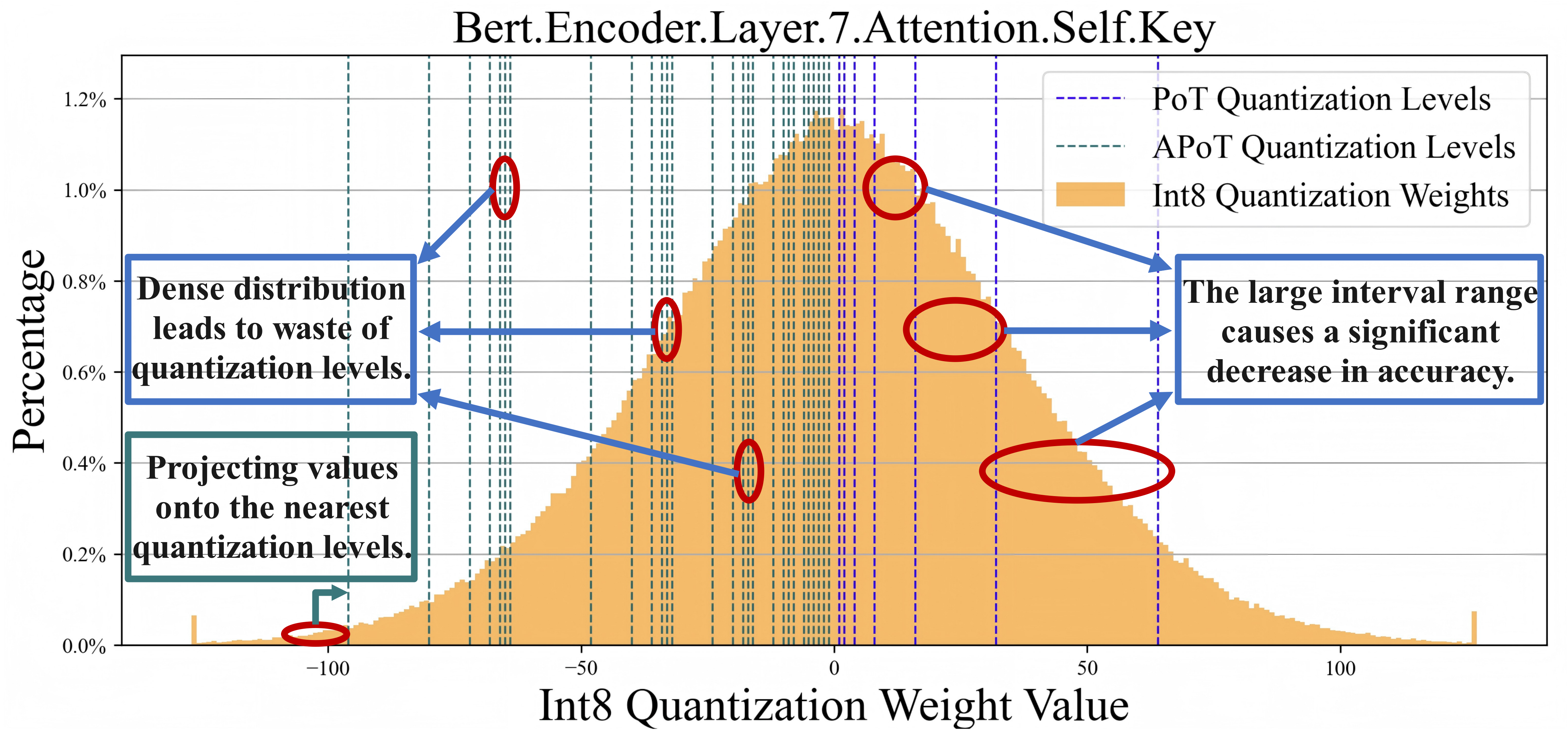}
  \caption{Distributions of 8-bit weights and quantization levels of PoT and APoT.}
  \label{fig:Quantization}
\end{figure}

To illustrate the limitations of the two quantization methods, 
we randomly select weights that have been quantized using 8-bit symmetric quantization. 
The distribution of these weights, along with the corresponding quantization levels of the two methods, 
is shown in Figure~\ref{fig:Quantization}.
Since PoT projects values to the one-hot encoding corresponding to the first leading one, 
the quantization level intervals increase as the data increases, 
resulting in unacceptable projection errors. 
In contrast, APoT projects values onto the sum of one-hot encodings 
corresponding to the closest a leading ones (with $a=2$ as an example). 
Although APoT improves precision by combining two one-hot vectors, 
the resulting dense quantization levels introduce a large number of redundant projection comparisons.

To address the above issues, we propose HLog,  
a hybrid of powers of two and their intermediate averages,
as illustrated in \eqref{eq:1}, where \(n\) is the bit-width. 
\begin{equation}
  \label{eq:1}
  \{ 2^0, 2^1, 2^0+2^1, 2^2, \ldots, 2^{n-2}, 2^{n-3}+2^{n-2}, 2^{n-1} \}
\end{equation}
If the data is equidistant from two adjacent quantization levels, it is projected to the higher quantization level.

\begin{figure}[t]
  \centering
  \includegraphics[width=\linewidth]{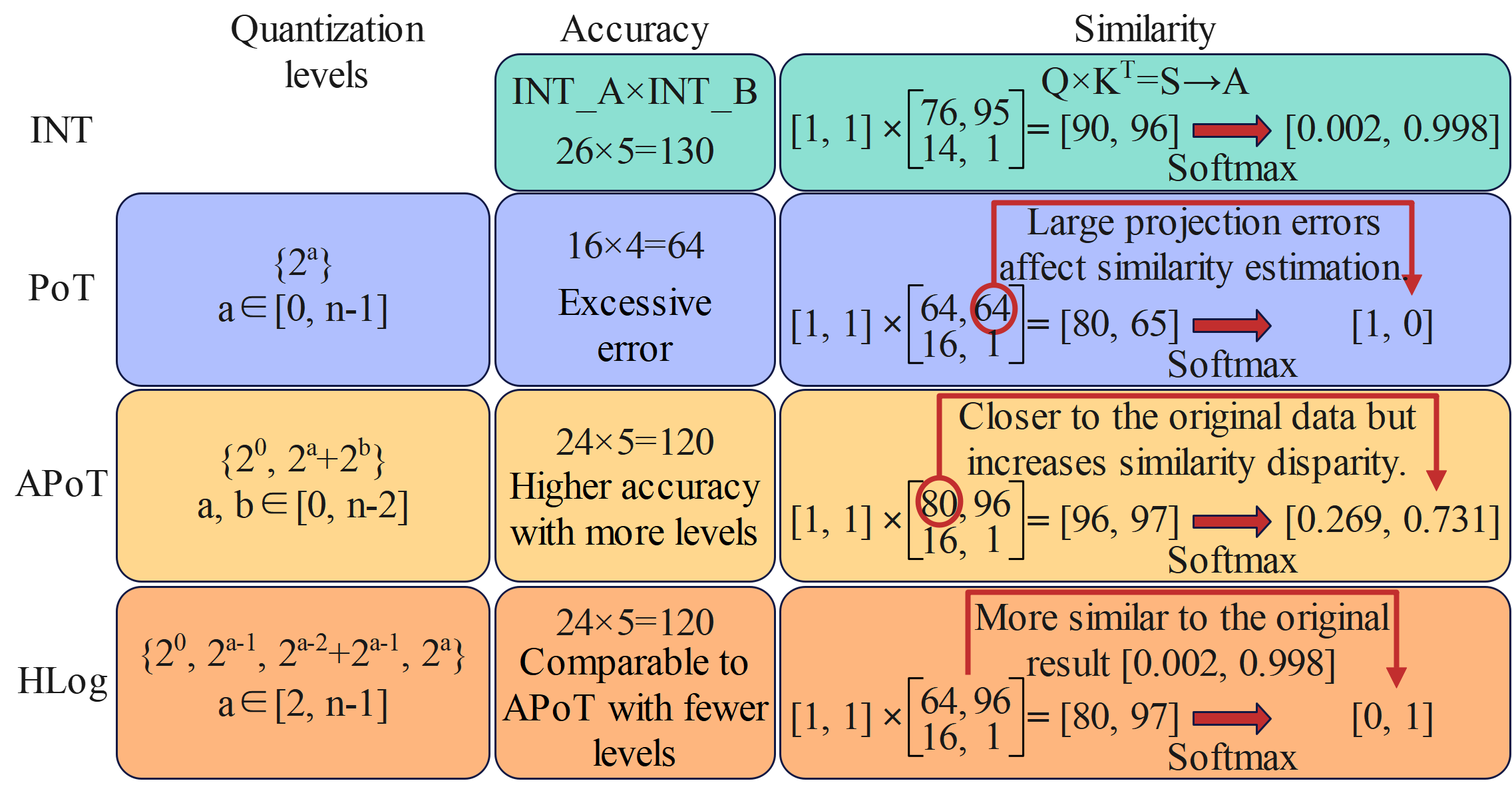}
  \caption{Comparison of three quantization methods.}
  \label{fig:quant_method}
\end{figure}

We illustrate the advantages of HLog from two perspectives: accuracy and similarity, 
as shown in Figure~\ref{fig:quant_method}.
Due to the limited number of quantization levels, PoT introduces large projection errors compared with APoT and HLog. 
These errors accumulate from the QK prediction to the attention prediction stage, 
leading to significant accuracy degradation.
Although APoT improves quantization accuracy by incorporating more quantization levels, 
the increased number of levels not only results in higher projection overhead but also degrades subsequent similarity prediction.
When the input values are relatively large, APoT tends to amplify non-maximum elements, whereas HLog conservatively reduces them.
Given the exponential nature of the softmax function, 
a slight reduction in non-maximum values better preserves the original probability distribution than a slight increase.
Moreover, the accumulation in vector multiplication further amplifies this numerical expansion, 
causing non-maximum elements to challenge the dominance of the maximum,
which distorts the original distribution and degrades similarity estimation.
Therefore, compared with PoT and APoT, 
HLog provides data that better preserve the original distribution for subsequent inter-row similarity estimation.

\subsection{Local Similarity Calculation}

To reduce the overhead of similarity computation, we perform similarity calculations on the SPA. 
To further enhance efficiency and leverage local similarity, we adopt a fixed-window strategy. 
Specifically, the full SPA of size \( L \times L \) is partitioned into non-overlapping windows of size \( w \times L \). 
In cases where \(L\) is not divisible by \(w\), 
the remaining rows are grouped into an additional window to ensure complete coverage.
Since each window can be processed independently, the similarity computation can be parallelized across windows, 
effectively reducing the latency introduced by similarity computation. 
We use the L1 distance to measure similarity among row vectors within a local window, 
thereby further reducing the cost of similarity computation.
The total computational cost for similarity estimation is $L^2 (w - 1)$ additions and subtractions.
Based on the similarity results, we partition the vectors into critical vectors and similar vectors, 
while maintaining a mapping between their row indices to preserve correspondence.

\begin{figure}[t]
  \centering
  \includegraphics[width=\linewidth]{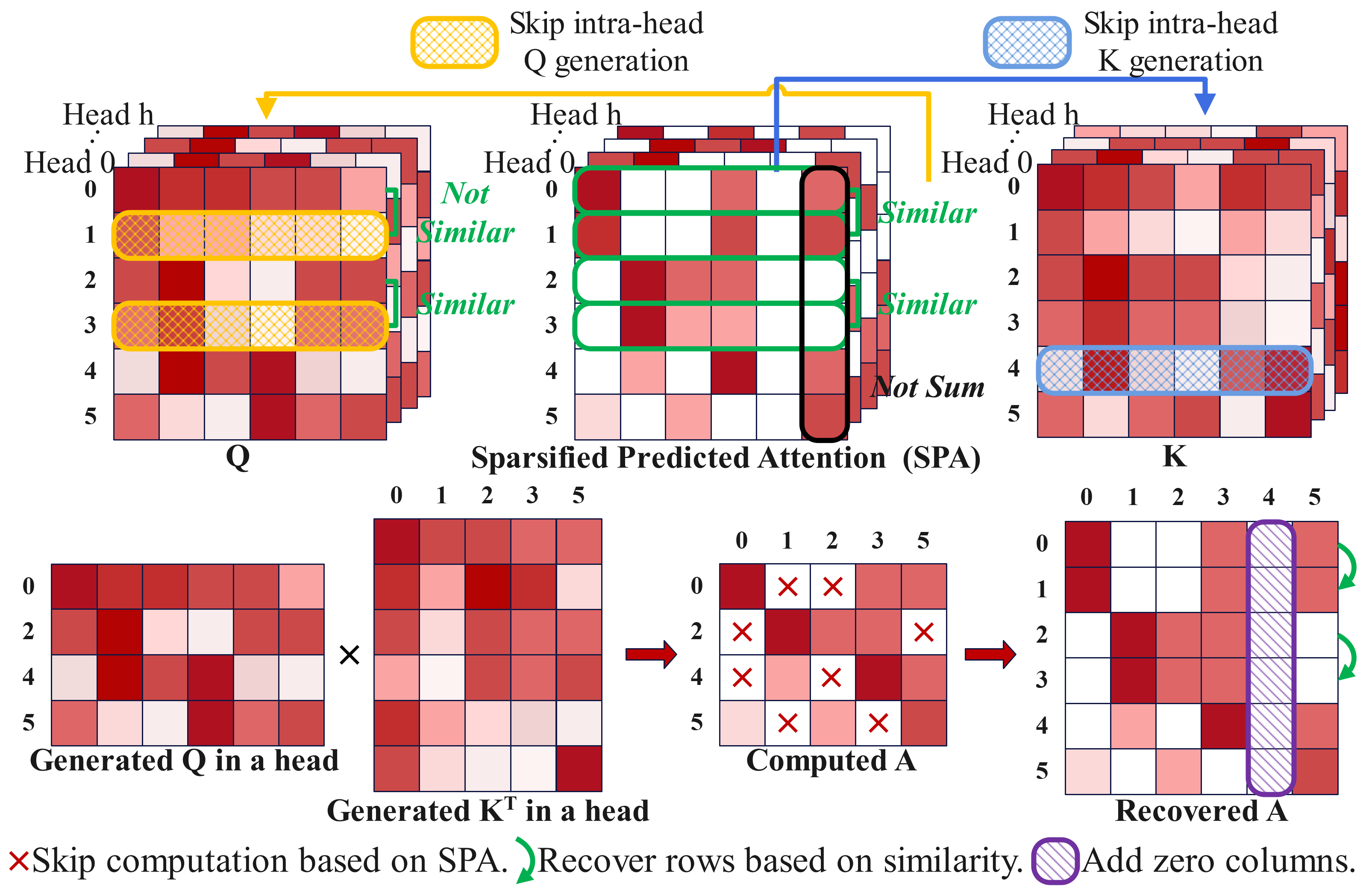}
  \caption{Overview of similarity-based Q and column-based K sparsification.}
  \label{fig:Q-SP}
\end{figure}

\subsection{Sparsification of QKV Generation}\label{sec:q_sp}

Due to the independence of QK computations across different heads in the multi-head attention mechanism, 
we apply sparsity to QKV separately for each head, 
based on the similarity and sparsity patterns observed in the corresponding SPA of that head.

\textbf{Similarity-based sparsification of Q.} 
As shown in Figure~\ref{fig:Q-SP}, performing similarity computation on the SPA, rather than on the dense predicted attention, 
leads to improved sparsity in Q. 
This results from the fact that, 
in the dense case, generating each attention row of length L requires all $L$ K vectors, 
which implies that only highly similar Q vectors can yield similar attention rows. 
In contrast, after applying Top-$k$ pruning, only \( L \times k \) K vectors are involved. 
As a result, even dissimilar Q vectors may produce attention rows that are similar in the key positions.
Moreover, this approach does not introduce additional system latency, 
as the similarity computation is merely postponed to occur after the Top-$k$ pruning. 
The overall latency remains determined solely by the similarity computation itself.

At the intra-head level, we sparsify the generation of Q.
For each head, Q vectors are generated only for the critical attention rows identified by similarity prediction.
Following the SPA, the corresponding QK multiplications in the attention matrix are skipped.
Since only critical attention rows are computed, 
a recovery operation is performed on the attention matrix by replicating the critical rows to their corresponding similar rows, 
thereby restoring the row dimension and maintaining the correct output shape.

\textbf{Column-based Sparsification of KV.}
To avoid introducing additional latency, we perform the sparsification of KV concurrently with the sparsity detection of Q. 
Instead of computing column-wise importance scores through summation, 
we directly identify zero columns in the SPA and use them as indicators to prune the corresponding rows in K. 
This is justified by the fact that the Top-$k$ operation has already removed unimportant elements, 
making further summation-based evaluation unnecessary. Moreover, 
detecting zero columns requires significantly fewer resources than summation, 
thereby reducing the computational overhead during prediction.
For V, since it is multiplied by the attention matrix in the subsequent stage, 
the rows in V that correspond to zero columns in attention are also redundant and can be safely pruned.

\subsection{Sparsification of FFN}

\begin{figure}[t]
  \centering
  \includegraphics[width=\linewidth]{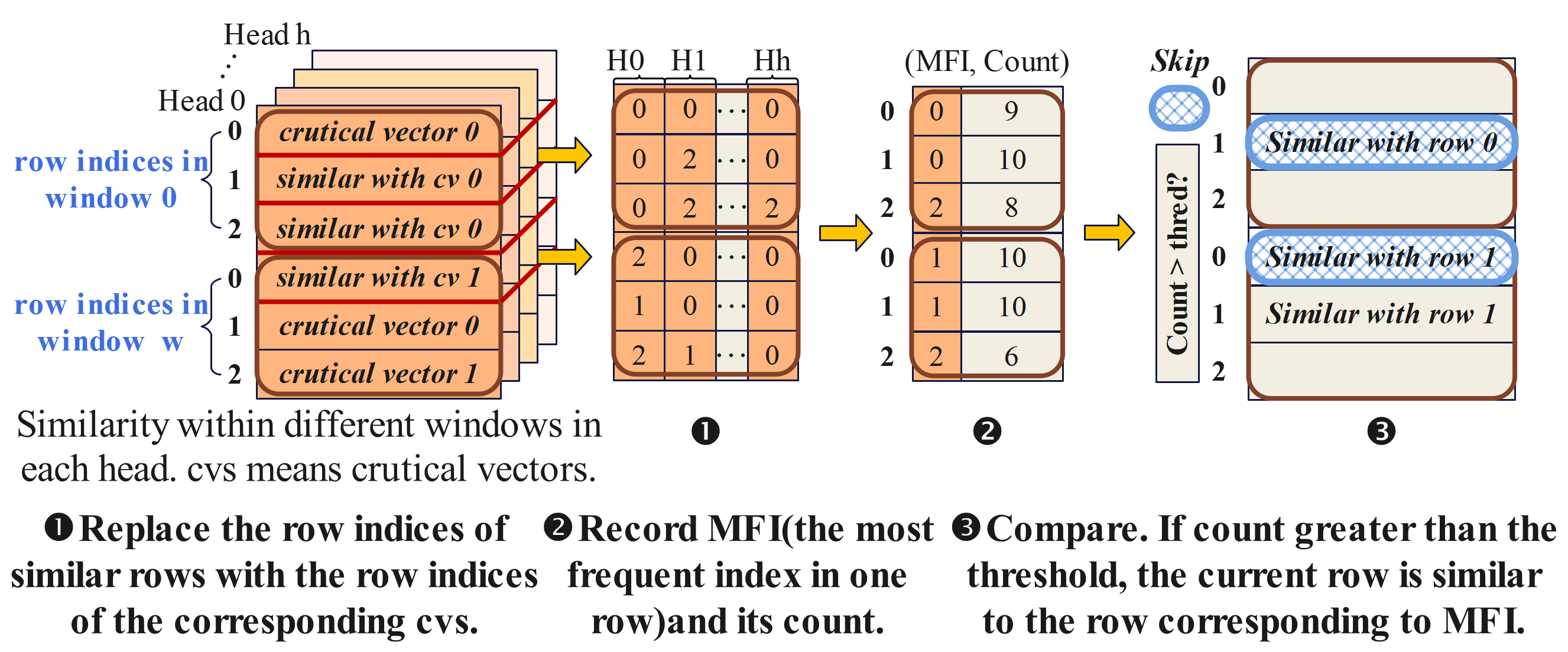}
  \caption{Sparsification of FFN with the most frequent index (MFI) method.}
  \label{fig:FFN-SP}
\end{figure}

We further leverage local similarity to induce sparsity in the FFN. Since the input to the FFN is the output of the MHA, 
a token may exhibit varying similarity patterns across different heads, 
which complicates the assessment of token-level similarity. 
To address this challenge, we propose the Most Frequent Index (MFI) method, 
which identifies token similarity 
by selecting the most frequently occurring critical vector across heads as the representative of each token, 
as shown in Figure~\ref{fig:FFN-SP}.
Specifically, during the concatenation of multi-head attention outputs, 
we first represent the attention result using the indices of the critical vectors. 
For each head, the row indices of similar vectors are replaced with those of their corresponding critical vectors. 
Next, for each token, we compute the most frequent critical index (MFI) and record its occurrence count. 
Finally, we determine whether the current token is similar to the token corresponding to the MFI 
by comparing the occurrence count with a predefined threshold.
Thanks to the previously applied fixed-window strategy, 
the above operations can be efficiently parallelized. 
Using the MFI method, we achieve token-level similarity assessment, 
which enables sparsification of the FFN. 
Similar to the sparsification of Q, 
a recovery step is performed after the final FFN layer to restore the original output shape.

\section{ESACT HARDWARE ARCHITECTURE}\label{sec:hardware}

\begin{figure}[t]
  \centering
  \includegraphics[width=\linewidth]{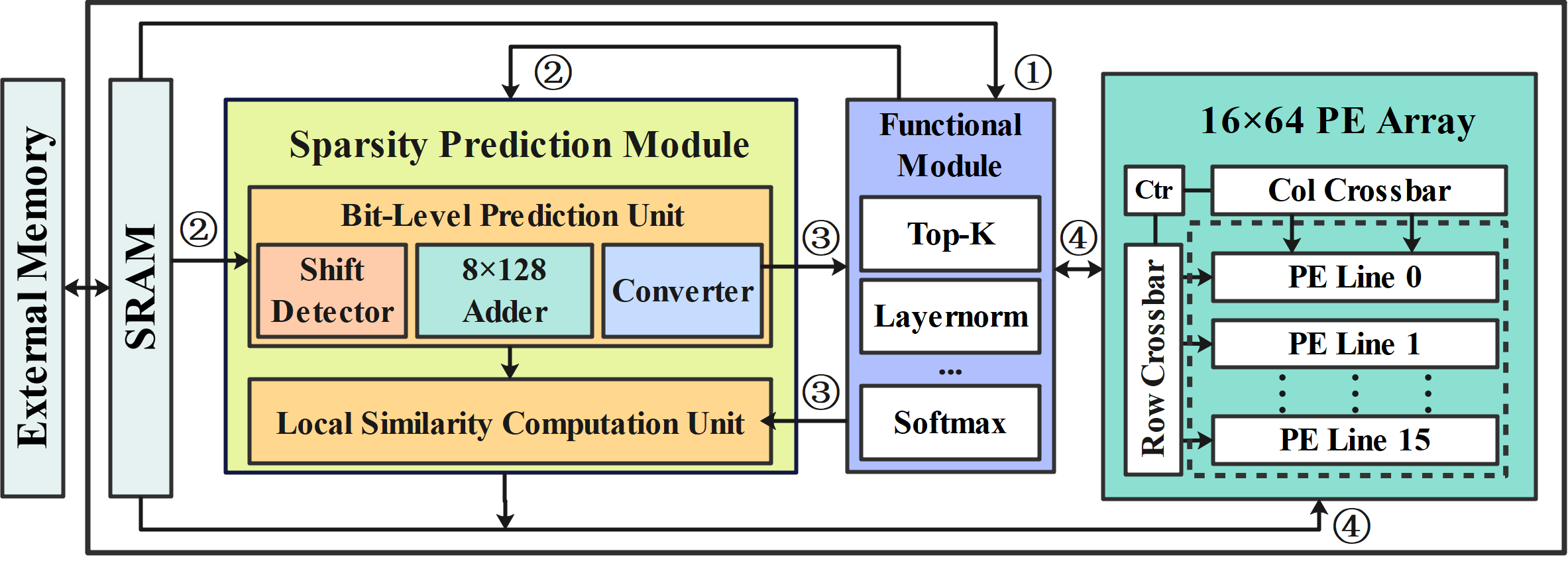}
  \caption{Overall architecture of ESACT accelerator.}
  \label{fig:hardware}
\end{figure}

\subsection{Overall Architecture}\label{sec:overall}

To better support the proposed algorithm, we introduce ESACT, 
an accelerator designed to exploit local similarity for sparse computation in QKV generation, 
attention computation, and the FFN. 
The overall architecture of ESACT is illustrated in Figure~\ref{fig:hardware}, comprising on-chip SRAM, 
a Sparsity Prediction Module, a PE array and a Functional Module.
\textcircled{1} After the input tokens are embedded via the Functional Module, 
ESACT processes each attention head independently. 
\textcircled{2} The embedding vectors and the linear transformation weights $W_Q$ and $W_K$ 
are jointly fed into the bit-level prediction unit within the Sparsity Prediction Module, 
where quantization and shift-add operations are performed to generate the PAM. 
\textcircled{3} The Top-$k$ operation within the Functional Module produces the SPA, 
which is then passed to the local similarity computation unit.
\textcircled{4} The PE array selectively performs QKV generation and attention computation 
based on the sparsity and similarity patterns predicted by the sparsity prediction module. 
By adopting a progressive generation scheme, sparsity prediction and QKV computation can be executed concurrently, 
thereby reducing PE array idle time.
Once attention computation across all heads is completed, 
a dynamic allocation strategy is employed during concatenation to generate the final attention output. 
This output is then used by the PE array to perform block-wise sparse FFN computation.

\begin{figure}[t]
  \centering
  \includegraphics[width=\linewidth]{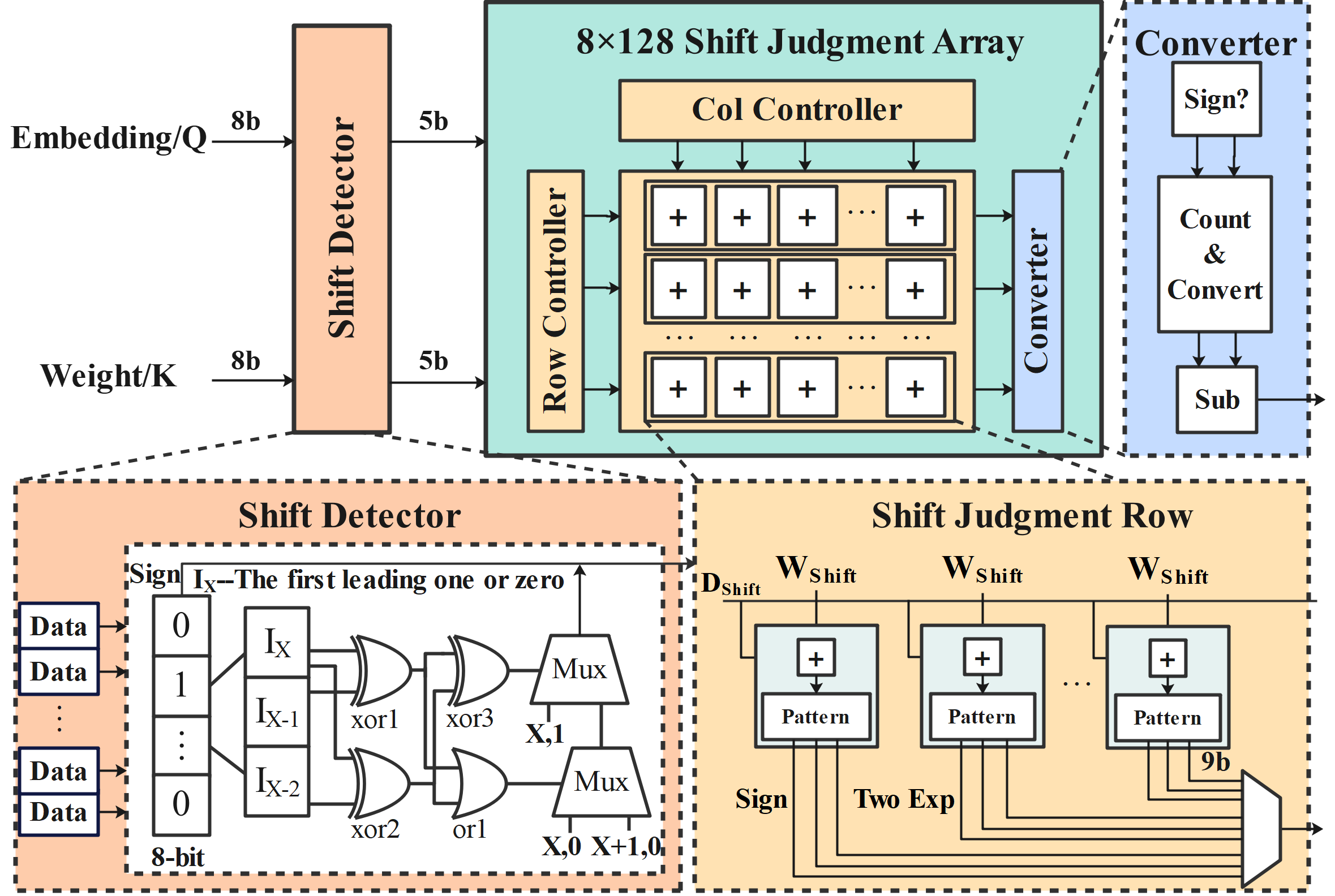}
  \caption{Components of the bit-level prediction unit.}
  \label{fig:hard_qu}
\end{figure}

\subsection{Bit-level Prediction Unit}

To enable efficient quantization and prediction, 
we design a bit-level prediction unit that leverages the correlation between adjacent bits and shift-add operations. 
As illustrated in Figure~\ref{fig:hard_qu}, the unit consists of three main components: a shift detector (SD), 
a shift judgment array (SJA) and a converter. 
Specifically, the 8-bit embedding and weight are first fed into the SD, where HLog quantization is performed. 
The quantized outputs are then forwarded to the SJA, 
where addition-only operations are employed to directly replace conventional multiplications, 
producing the intermediate products. 
These products are subsequently aggregated in the converter to generate the predicted QK. 
After obtaining the QK predictions, an additional 8-bit quantization is performed, 
and the entire process is repeated to predict the attention matrix.

\begin{figure}[t]
  \centering
  \includegraphics[width=\linewidth]{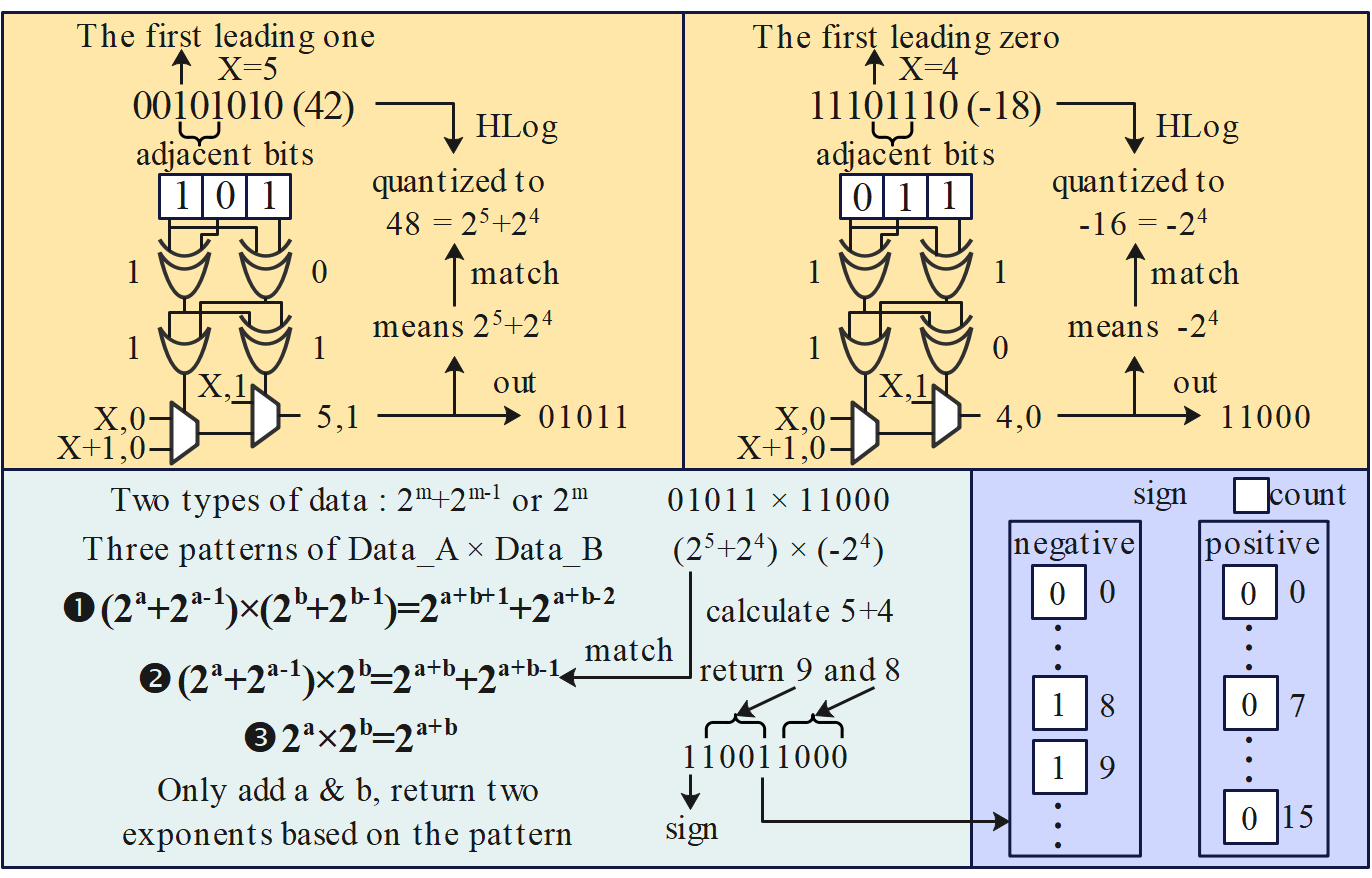}
  \caption{An example of the processing of data in the bit-level prediction unit.}
  \label{fig:hard_qu_example}
\end{figure}

To provide a more intuitive illustration of our design, 
we present a representative example using two 8-bit inputs: $(00101010)_2$ and $(11101110)_2$. 
As shown in Figure~\ref{fig:hard_qu_example}, when the inputs are fed into the SD, 
the unit first identifies the first leading one or zero based on the sign of the data, 
denoted as $I_x$. 
Subsequently, $I_x$ and the following two bits are extracted for quantization analysis. 
Specifically, the bit patterns $(101)_2$ and $(011)_2$ undergo xor and or operations, 
whose results are used to determine the corresponding quantization codes (5,1) and (4,0). 
Here, the first value indicates the exponent of the dominant power-of-two component, 
while the second value (0 or 1) specifies the quantization form: 
whether the result is represented by a single power-of-two or a sum of two powers-of-two.
The quantized results are then combined with the original sign bit to form a 5-bit output: 
$(01011)_2$ and $(11000)_2$, where the most significant bit represents the sign, 
the least significant bit indicates the quantization type (single or sum form), 
and the middle three bits encode the exponent of the largest power-of-two component. 
The SD enables the HLog quantization process to be performed without adding extra errors, 
achieving a bit-efficient representation of input values.

The quantized data are then fed into the SJA, 
where addition-based operations are performed according to their quantization formats. 
Due to the use of HLog quantization, each value is represented in one of two forms: either $2^m$ or $2^m+2^{m-1}$. 
Consequently, the multiplication of two quantized inputs can be categorized into three distinct cases, 
as illustrated in Figure~\ref{fig:hard_qu_example}.
By summing the exponents of the two inputs and using the least significant bit of the SD output to determine the multiplication mode, 
the SJA computes the product using only addition operations, entirely avoiding multipliers. 
This design significantly reduces the power consumption during prediction.
After obtaining the product, we encode the result by combining its sign bit with the two 4-bit exponents of the power-of-two terms, 
yielding a compact 9-bit output that preserves all necessary information for downstream processing.

Since the outputs of the SJA represent exponents of power-of-two terms, 
we design the converter by drawing inspiration from the one-hot adder architecture in FACT. 
The Converter first groups the inputs according to their sign bits, and then counts the occurrences of each exponent within each group. 
These counts are subsequently converted into binary representations, 
after which a subtraction is performed between the positive and negative groups to produce the final prediction result.

\begin{figure}[t]
  \centering
  \includegraphics[width=\linewidth]{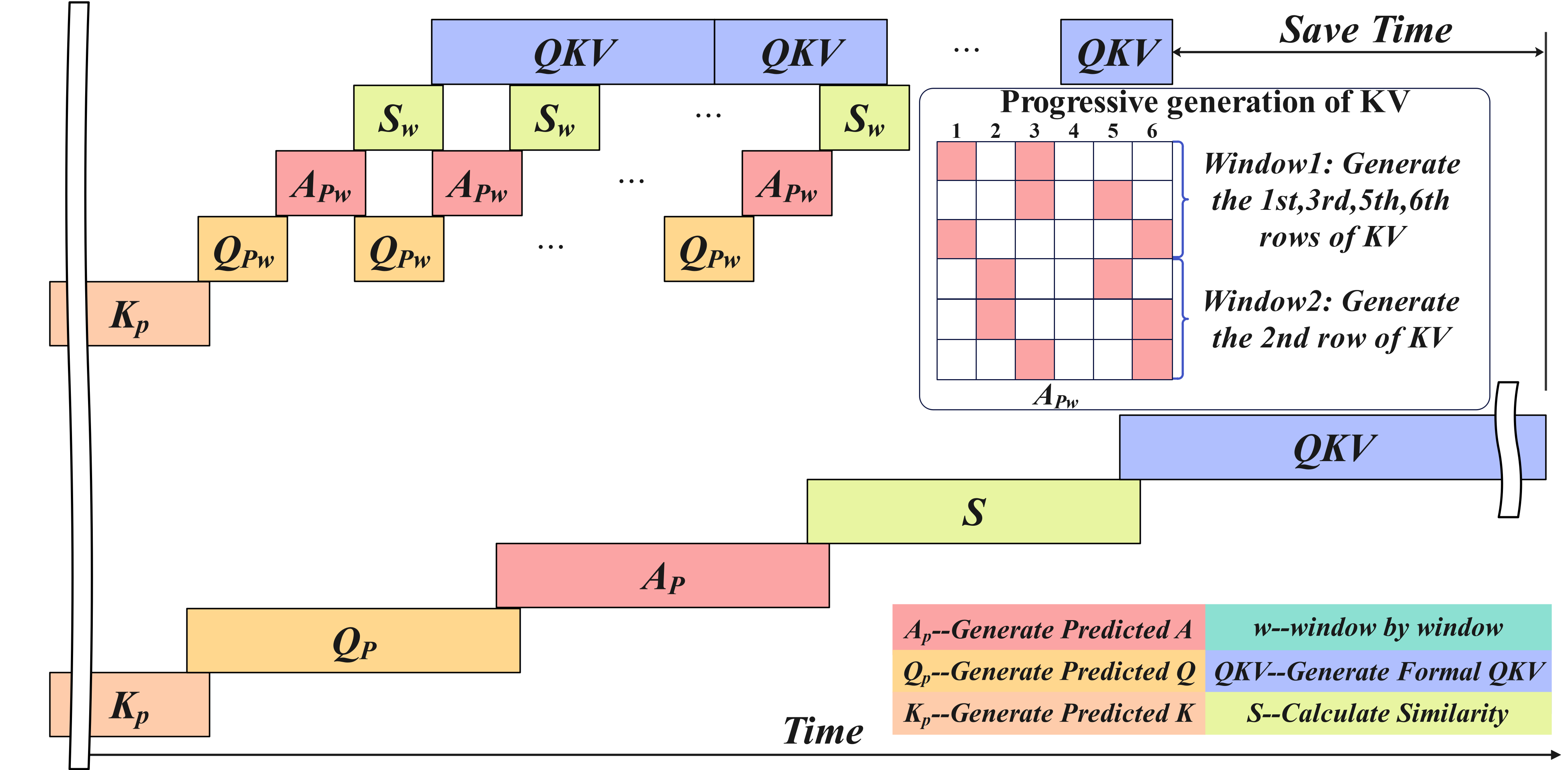}
  \caption{Progressive generation scheme.}
  \label{fig:progressive}
\end{figure}

\subsection{Progressive Generation Scheme}

\begin{figure*}[t]
  \centering
  \includegraphics[width=\linewidth]{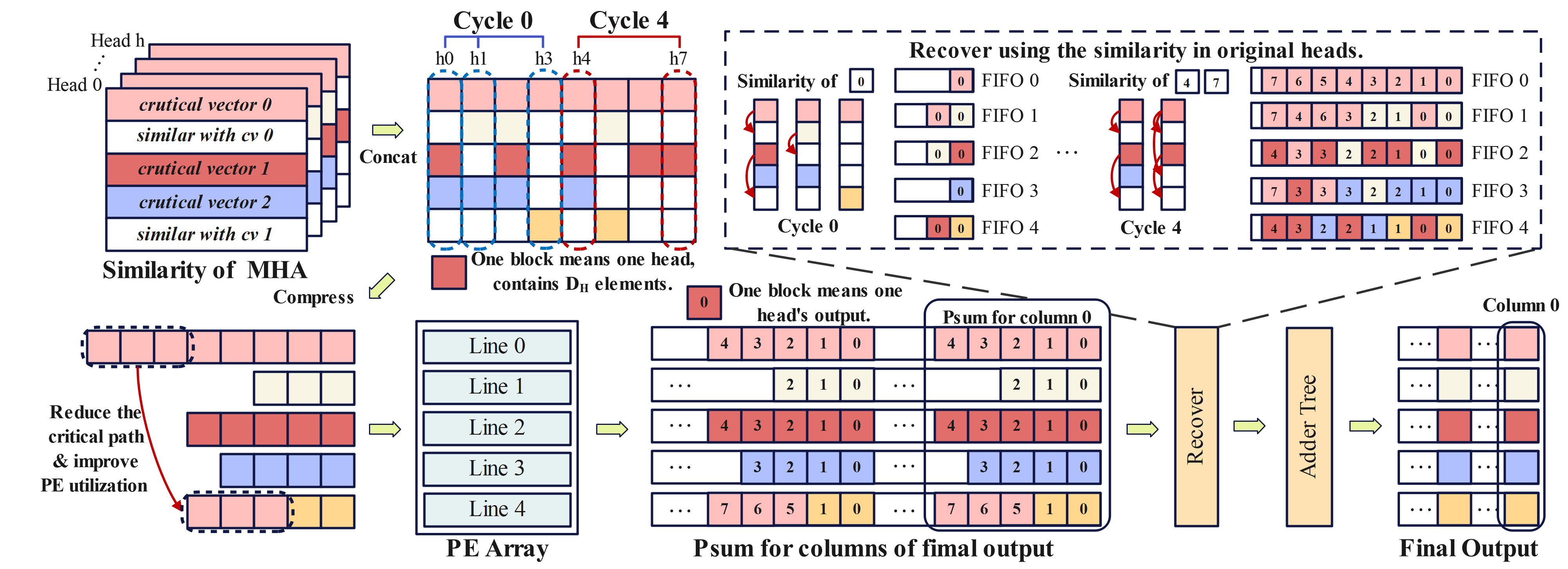}
  \caption{Dynamic allocation strategy.}
  \label{fig:FFN-ba}
\end{figure*}

Prior to the formal QKV generation, our prediction algorithm involves three sequential stages: 
QK prediction, attention prediction, and similarity computation, which introduces additional system latency. 
To mitigate this issue, we propose a progressive generation scheme, as illustrated in Figure~\Ref{fig:progressive}.
The scheme begins by predicting all K vectors. 
Given that we employ a fixed-window similarity strategy, the computations across different windows are independent. 
This allows per-window prediction of Q, attention, and similarity in a sequential yet parallelizable manner. 
Once the sparsity and similarity results for a given window are obtained, QKV generation can proceed progressively.
For Q, progressive generation is performed window by window, based on the similarity relationships identified within each window. 
For KV, the generation process is guided by the distribution of empty columns in each window.
In the figure, the $2nd$ and $4th$ columns in window$1$ are identified as empty, 
so only the $1st$, $3rd$, $5th$ and $6th$ rows of the KV matrix are generated. 
Since the $2nd$ column becomes active in window$2$, the $2nd$ row of KV is generated at that stage. 
Through this progressive and conditional KV generation, 
we effectively eliminate redundant computation that would otherwise occur if each window generated KV independently.

By performing prediction and QKV generation in a progressive and window-wise manner, 
the scheme enables overlap between the prediction and formal generation phases, 
thereby reducing the idle time of PE array and significantly lowering overall system latency.

\subsection{Dynamic Allocation Strategy}

Due to the multi-head attention mechanism, the similarity distribution varies across different heads, 
resulting in irregular sparsity patterns after attention concatenation. 
This irregularity poses challenges for the PE array, which is unable to efficiently leverage sparsity. 
To address this issue, we propose a dynamic allocation strategy, as illustrated in Figure~\ref{fig:FFN-ba}.
After concatenation, the number of preserved critical vectors differs from row to row, 
causing load imbalance across PE lines. 
To mitigate this, 
we first apply compression to the concatenated attention map and then perform dynamic matching based on the compressed data distribution, 
which shortens the critical path and improves PE utilization.
The PE array adopts a weight-stationary execution model, 
where weights are assigned based on the input and held stationary during computation to generate the output for each head, 
represented as partial sum (Psum). Since only the Psum corresponding to critical vectors are computed explicitly, 
a recovery mechanism based on intra-head similarity is employed to reconstruct the missing Psum for similar vectors, 
thereby producing the final output.
As shown in the figure, once the first group of heads in each PE Line completes computation and generates the Psum represented 
by block$0$, 
we utilize the intra-head similarity of block$0$s within head$0$, head$1$ and head$3$ to assign values to the corresponding FIFOs. 
This process is referred to as cycle$0$. 
In cycle$4$, the intra-head similarity within head$4$ and head$7$ is similarly used to perform the recovery.
Eventually, each FIFO accumulates eight Psums, which are then summed to produce the final output.

Through the dynamic allocation strategy, we effectively leverage the sparsity induced by similarity to reduce overall computation. 
Additionally, by shortening the critical path, the strategy mitigates load imbalance caused by irregular sparsity patterns, 
thereby improving the utilization of the PE array.

\section{EVALUATION}\label{sec:eval}
\subsection{Workloads}

We evaluate the performance of ESACT across 26 benchmarks. 
For NLP tasks, we adopt BERT-Base and BERT-Large~\cite{bert} as representative transformer encoders. 
To assess ESACT's effectiveness under varying sequence lengths, 
we select eight tasks from the GLUE~\cite{glue} benchmark (excluding WNLI), along with SQuAD v1.1~\cite{squad} and CLOTH~\cite{cloth}, 
corresponding to sequence lengths of 128, 384, and 512, respectively. 
To evaluate ESACT's applicability to decoder-based models, we further assess GPT-2~\cite{gpt2}, 
Llama2-7b~\cite{llama} and Bloom-7b~\cite{bloom} on the WikiText-2~\cite{wiki} dataset.
For the CV domain, we evaluate ViT-B/16 and ViT-B/32~\cite{vit} on the ImageNet-1K~\cite{imagenet} dataset. 
All models are implemented using PyTorch~\cite{pytorch} and the Huggingface Transformers Library~\cite{huggingface}. 
As the models are pre-trained, we fine-tune them on the respective datasets.
We use a batch size of 32 for GLUE tasks, 12 for SQuAD, 3 for CLOTH, and 8 for both WikiText-2 and ImageNet-1K. 
The learning rate is set to $2 \times 10^{-5}$ for GLUE and ImageNet-1K, $3 \times 10^{-5}$ for SQuAD, 
and $5 \times 10^{-5}$ for both CLOTH and WikiText-2.

\begin{figure*}[t]
  \centering
  \includegraphics[width=\linewidth]{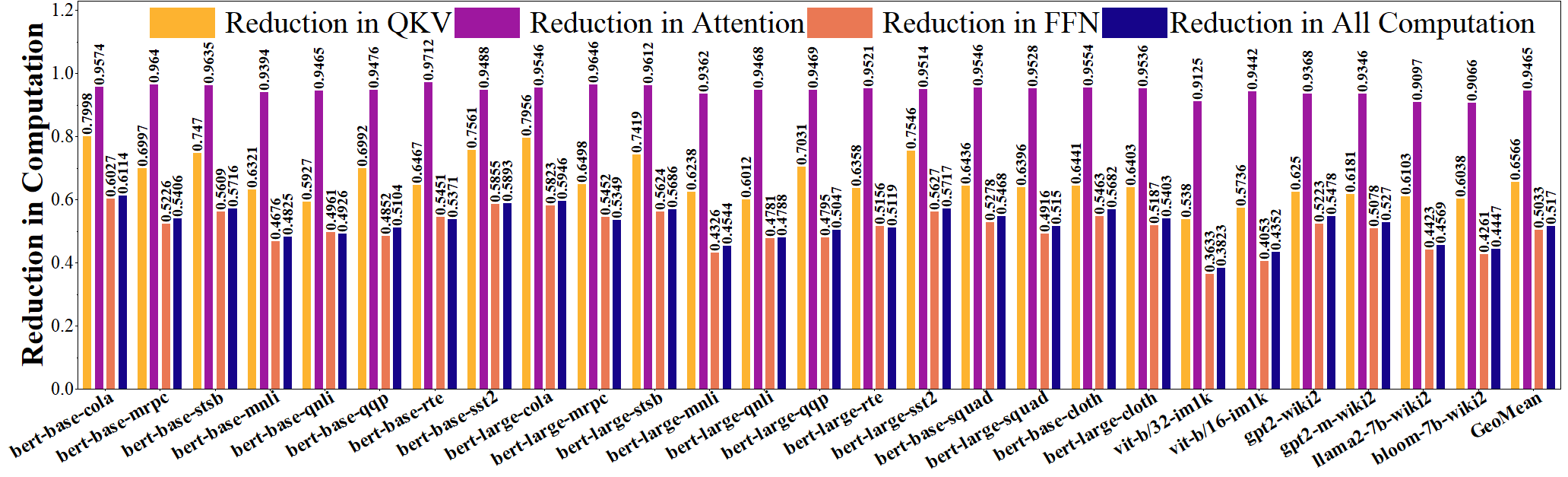}
  \caption{Overall computation reduction and component-wise breakdown under the loss $\leq 1\%$.}
  \label{fig:reduction_and_acc}
\end{figure*}

\begin{figure*}[t]
  \centering
  \includegraphics[width=\linewidth]{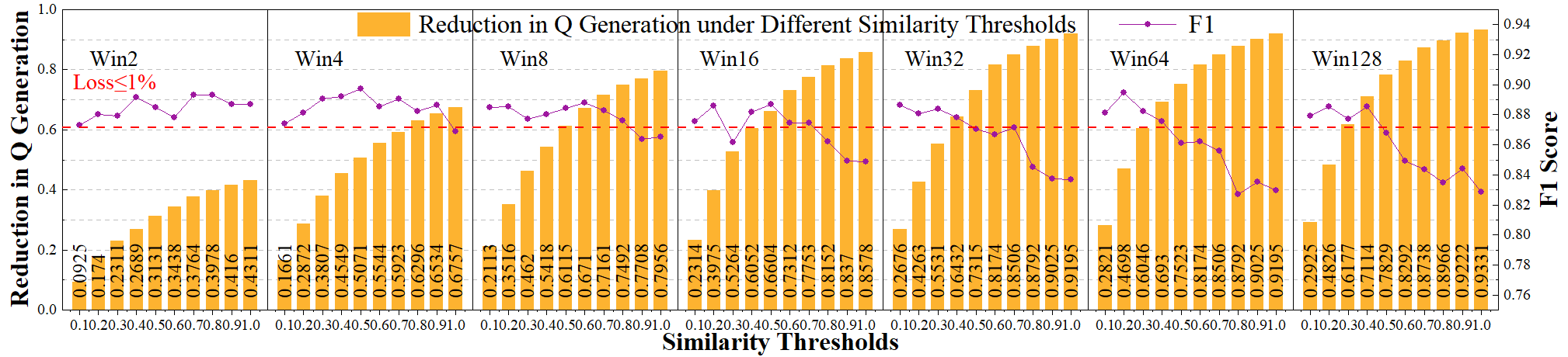}
  \caption{Impact of similarity threshold $s$ and window size on Q sparsity and model accuracy on the MRPC task. 
  Results are based on fine-tuning BERT-Base with a fixed top-$k$ ratio of 0.12, without applying FFN sparsification.}
  \label{fig:reduction_of_win}
\end{figure*}

\subsection{Accuracy Evaluation}
\textbf{Methodology.} 
We begin by applying 8-bit quantization-aware fine-tuning to model weights on eight Nvidia V100 GPUs
across all tasks, 
maintaining accuracy while enabling efficient deployment on hardware platforms.
The SPLS mechanism has three important hyperparameters: the top-$k$ ratio $k$, 
the similarity threshold $s$, and the FFN threshold $f$. 
These hyperparameters regulate sparsity in different components: 
smaller $k$ for Attention, larger $s$ for QKV, and smaller $f$ for FFN induce greater sparsity in their respective modules.
Starting from the quantized models, we first apply top-$k$ pruning and determine the optimal $k$ value that preserves task accuracy.
With this $k$ fixed, we perform a fine-grained grid search over the $(s, f)$ space, 
varying $s$ with a step size of 0.02 and $f$ with a step size of 1. 
For each task, we fine-tune the model under different $(s, f)$ configurations 
and retain those in which the performance degradation remains within acceptable bounds (i.e., loss $\leq 1\%$).
Task-specific evaluation metrics are used to assess performance: F1 score for MRPC, QQP, and SQuAD; 
accuracy for SST-2, QNLI, MNLI, RTE, ImageNet-1K, and CLOTH; and perplexity for WikiText-2.
We consider a configuration acceptable if the drop in F1 score or accuracy is no more than 1\%, 
or if the increase in perplexity remains below 5\%.

\textbf{Overall Computation Reduction and Component-wise Breakdown.} 
As illustrated in Figure~\ref{fig:reduction_and_acc}, under a local window size of 8, 
we present the overall computation reduction as well as the reductions in QKV, 
attention, and FFN computations across all tasks where the loss remains within 1\%. 
On average, the proposed SPLS mechanism achieves a 51.7\% reduction in overall computation, 
with 65.66\%, 94.65\%, and 50.33\% reductions in QKV, attention, and FFN computations, respectively.
The significant reduction in attention computation is attributed to both 
\textit{inter-row sparsity} enabled by local similarity and \textit{intra-row sparsity} from top-$k$ pruning. 
These results demonstrate the effectiveness and broad applicability of the SPLS mechanism, 
primarily because all evaluated models generate attention in a fully parallel manner during fine-tuning, 
which is well aligned with the assumptions underlying our method.

\textbf{Impact of Similarity Threshold and Window Size.} 
To investigate the impact of the similarity threshold $s$ and window size in the SPLS mechanism on both accuracy and sparsity, 
we fine-tune BERT-Base on the MRPC task under a fixed top-$k$ ratio of 0.12.
We vary the similarity threshold $s$ from 0.1 to 1.0 and 
examine its effect on model accuracy and the sparsity of the Q generation across different local window sizes.
The results are presented in Figure~\ref{fig:reduction_of_win}.
As shown in the figure, increasing $s$ for a given window size consistently yields higher sparsity in the Q generation.
Interestingly, model accuracy remains stable or even slightly improves across a range of $s$ values before eventually degrading,
suggesting that local similarity patterns exist in the input and can be exploited to improve attention quality.
By selecting representative tokens based on local similarity, 
the model can reduce redundant computation while maintaining or enhancing performance.
When varying the window size, we observe a consistent overall trend, 
but smaller windows (e.g., 2 or 4) fail to induce substantial sparsity.
This is likely due to the limited token context within small windows, 
which hampers the ability to capture meaningful similarity.
On the other hand, excessively large window sizes incur higher computational overhead. 
To strike a balance between sparsity and efficiency, we adopt a window size of 8 for all experiments in this work.

\begin{figure}[t]
  \centering
  \includegraphics[width=\linewidth]{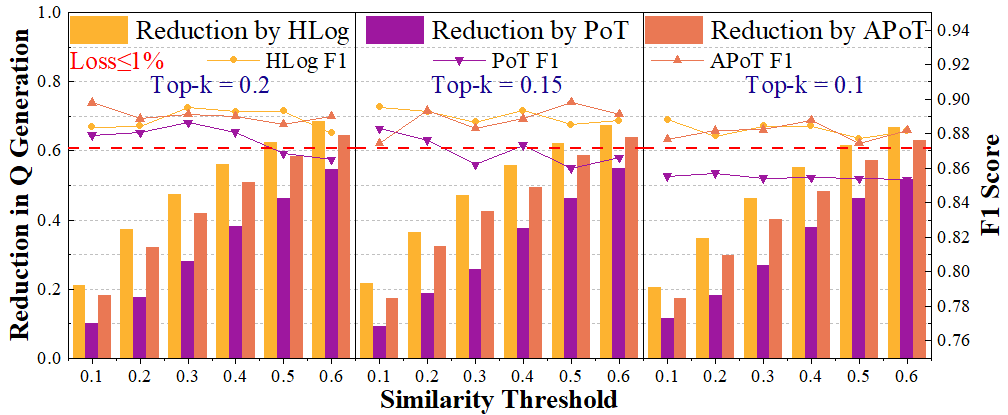}
  \caption{Comparison of Q sparsity and model accuracy under different quantization methods (HLog, PoT, and APoT) on BERT-Base MRPC.}
  \label{fig:reduction_quant}
\end{figure}

\begin{figure}[t]
  \centering
  \includegraphics[width=\linewidth]{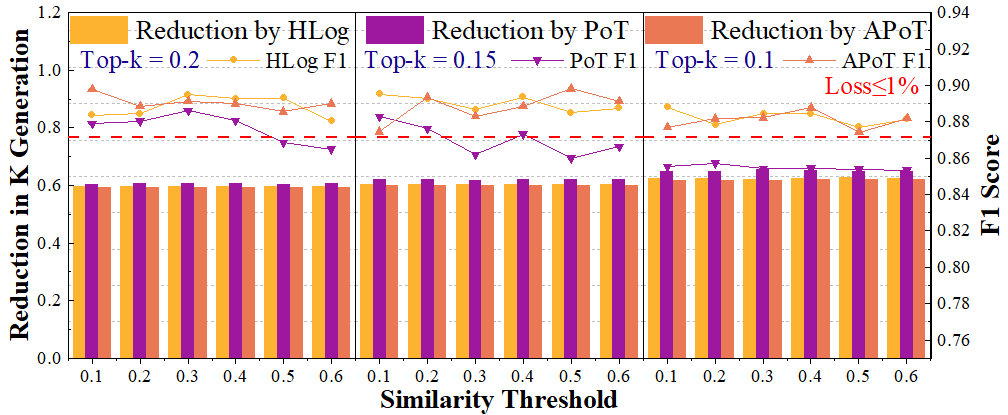}
  \caption{Comparison of K sparsity under different quantization methods (HLog, PoT, and APoT) on BERT-Base MRPC.}
  \label{fig:reduction_k}
\end{figure}

\textbf{Impact of Quantization Methods.} 
Before applying local similarity, we first employ HLog Quantization to predict the attention matrix, 
producing the Predicted Attention matrix (PAM). 
Row-wise top-$k$ pruning is then performed to the PAM to obtain the Sparsified Predicted Attention (SPA), 
which serves as the input for subsequent local similarity computation.
Since the quantization scheme directly affects the quality of similarity estimation, 
we compare three methods—HLog, PoT, and APoT—in terms of their impact on Q sparsity and model accuracy.
As shown in Figure~\ref{fig:reduction_quant}, 
HLog consistently achieves a more favorable sparsity-accuracy trade-off compared to both PoT and APoT.
Compared to PoT, HLog yields higher Q sparsity and better model accuracy under the same $k$ and $s$, 
with only a marginal increase in the number of quantization levels. 
In comparison to APoT, 
HLog provides even higher sparsity at the same level of accuracy and further eliminates redundant quantization levels.
This improvement is attributed to HLog's quantization level distribution, 
which more closely matches the distribution of the original weights. 
As a result, PAM better preserves the similarity structure of the true attention, 
enabling more effective pruning under fixed hyperparameters.

The sparsity of K under different quantization methods is shown in Figure~\ref{fig:reduction_k}. 
It can be observed that, under the same quantization prediction, the sparsity of K remains unchanged with respect to $s$, 
since it is determined by the empty columns generated by the Top-$k$ selection and is independent of inter-row similarity. 
For the same $k$ value, both APoT and HLog produce nearly identical sparsity with comparable precision, 
indicating the redundancy of APoT's quantization levels. 
In contrast, PoT quantization introduces large projection errors, 
causing some critical values to be discarded due to low prediction accuracy, 
which leads to a significant degradation in overall accuracy.
In summary, the experimental results demonstrate that 
our HLog outperforms both APoT and PoT in terms of similarity prediction and accuracy estimation.

\begin{figure}[t]
  \centering
  \includegraphics[width=\linewidth]{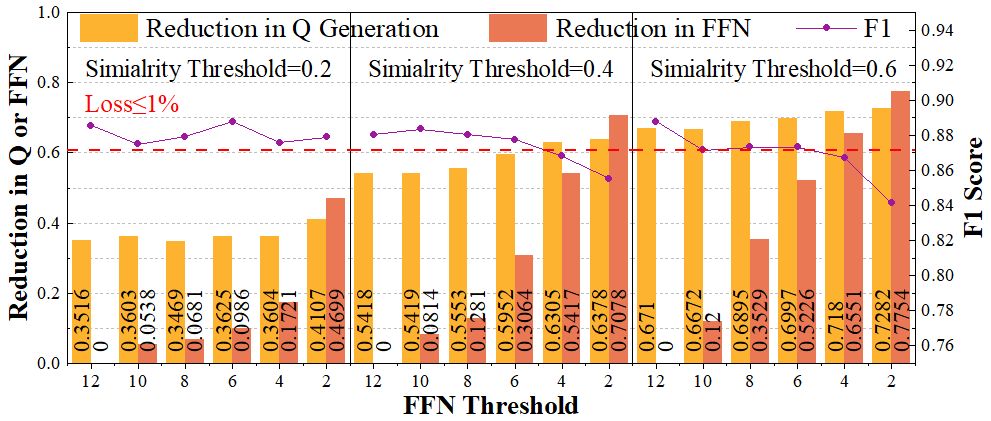}
  \caption{Impact of FFN threshold $f$ on Q and FFN sparsity and model accuracy on MRPC.}
  \label{fig:reduction_ffn}
\end{figure}

\textbf{Impact of FFN Threshold.} 
To investigate the impact of the FFN threshold $f$ on both sparsity and accuracy,
we fine-tune BERT-Base on the MRPC task and 
evaluate how varying $f$ affects model performance and sparsity under different $s$.
The results are shown in Figure~\ref{fig:reduction_ffn}.
We observe that the Q sparsity remains largely unaffected as the FFN threshold decreases,
indicating that the sparsity introduced in the FFN of the current layer 
does not propagate to the Q generation in the subsequent layer.
This decoupling is primarily due to architectural elements such as residual connections,
which disrupt the continuity of similarity patterns across layers and 
allow Q and FFN sparsity to be optimized independently.
As the FFN threshold $f$ decreases, 
the requirement for each token to retain a larger number of critical vector dimensions becomes less stringent.
This leads to greater inter-token similarity and, consequently, higher sparsity in the FFN layer.

\subsection{Performance Evaluation}
\textbf{Methodology.} 
To evaluate the performance of ESACT, we use Verilator~\cite{verilator} to simulate RTL and obtain the cycle counts of each stage. 
Based on these results, we build a custom cycle-level simulator that models the overall system execution. 
The baseline cycle counts obtained from Verilator are measured under a representative workload 
(sequence length $L = 128$, embedding dimension $D = 768$), 
with stage-specific sparsity: Q, K, and V each with 60\% sparsity, 
attention 60\% inter-row sparsity with intra-row top-$k$ values of 0.1, 0.15, and 0.2, and FFN 50\%. 
In the cycle-level simulator, per-stage latencies are computed using scaling functions---QKV, attention inter-row, and FFN, 
being structurally sparse, 
are scaled according to their sparsity ratios as well as $L$ and $D$, 
while attention intra-row cycles are predicted using a fitted function over the measured top-k points. 
For memory accesses, Ramulator~\cite{ramulator} is employed to model the off-chip DRAM timing behavior, including both latency and bandwidth effects. 

As the baseline, we measure the throughput of each model after 8-bit quantization-aware fine-tuning, 
executed on an Nvidia V100 GPU (32GB) with an Intel Xeon Gold 6130 CPU across multiple tasks.
For a fair comparison, we follow the setup in ~\cite{fact,elsa} by simulating 125 parallel ESACT units running at 500\,MHz, 
resulting in a theoretical peak throughput of 125\,TOPS, the same as that of the V100 GPU.
We divided the 125 ESACTs into 25 clusters. 
Each workload is partitioned along the batch, head, and sequence length dimensions, 
and then assigned to the clusters in order from the lowest to the highest dimension for parallel computation.
To match the V100, memory bandwidth is also configured to 900\,GB/s, 
ensuring that performance differences are not attributed to memory access limitations.
Based on accurate cycle-level simulation, a single ESACT requires a maximum bandwidth of 4.7\,GB/s, 
observed under intra-row sparsity for attention at the lowest PE utilization, 
specifically at $k = 0.1$ with a PE utilization of 81.57\%. 
Thus, memory bandwidth does not constitute a performance bottleneck, 
and the computation latency is dominated by compute rather than memory transfer.

\begin{figure}[t]
  \centering
  \includegraphics[width=\linewidth]{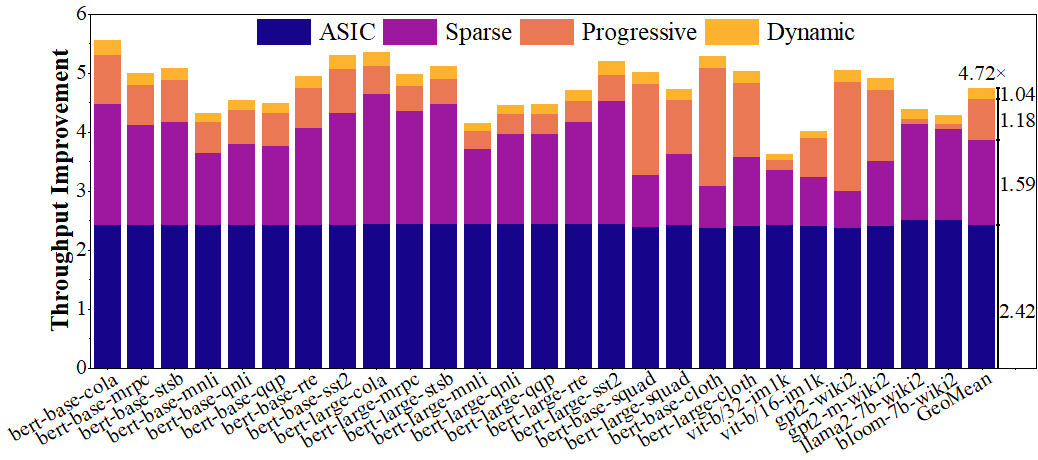}
  \caption{Breakdown of end-to-end throughput improvement of ESACT over Nvidia V100.}
  \label{fig:speed}
\end{figure}

\textbf{Throughput Improvement.} 
Figure~\ref{fig:speed} shows the end-to-end throughput improvements achieved by ESACT compared to the Nvidia V100 GPU. 
On average, ESACT delivers a $4.72\times$ speedup in end-to-end inference. 
This performance gain stems from a combination of architectural and algorithmic optimizations. 
The underlying ASIC design represents a dense version of the ESACT accelerator, 
employing a weight-stationary dataflow and sustaining 100\% of the peak throughput. 
It achieves a $2.42\times$ speedup over the V100, serving as the ideal baseline for performance comparison.
On top of this, our SPLS mechanism significantly reduces the overall computation by more than half, 
resulting in an additional $1.59\times$ improvement. 
To further enhance efficiency, 
the \textit{progressive generation scheme} overlaps formal QKV computation with sparsity prediction, 
effectively reducing PE idle time and contributing another $1.18\times$ speedup. 
Finally, the \textit{dynamic allocation strategy} alleviates critical path bottlenecks, 
yielding a further $1.04\times$ improvement in throughput.

\subsection{Area and Energy Evaluation}
\textbf{Methodology.} 
We implement the logic components of ESACT in Verilog and generate the SRAM blocks using ARM Memory Compiler. 
The entire design is synthesized using Synopsys Design Compiler with the TSMC 28nm standard cell library, 
targeting a clock frequency of 500\,MHz to obtain area and power reports.
For logic synthesis, we set the upper bound of the top-$k$ ratio to 0.2 
in order to reduce the number of subtractors required in the local similarity computation unit.

\begin{table}[t]
  \renewcommand{\arraystretch}{1.5} 
  \centering
  \caption{Area and power breakdown of ESACT at 500MHz}
  \label{tab:area}
  \resizebox{\columnwidth}{!}{%
    \begin{tabular}{|c|c|c|c|}
      \hline
      \textbf{Module} & \textbf{Parameter} & \textbf{Area (mm\textsuperscript{2})} & \textbf{Power (mW)} \\
      \hline
      PE Array & 
      \parbox{3cm}{\vspace{0.5em}\centering 16$\times$64 PEs\vspace{0.5em}} & 
      1.85 & 
      324.14 \\
      \hline
      Sparsity Prediction Module & 
      \parbox{3cm}{\vspace{0.5em}\centering 8$\times$26 Subtractors \\ 128 Shift Detectors \\ 8$\times$128 Adders \\ Converter\vspace{0.5em}} & 
      0.23 & 
      57.43 \\
      \hline
      SRAM & 
      \parbox{3cm}{\vspace{0.5em}\centering 192KB weight buffer \\ 192KB token buffer \\ 128KB temp buffer\vspace{0.5em}} & 
      1.6 & 
      317.84 \\
      \hline
      Functional Module & 
      \parbox{3cm}{\vspace{0.5em}\centering Top-k \\ LayerNorm \\ Softmax \\ others\vspace{0.5em}} & 
      1.41 & 
      92.71 \\
      \hline
      \multicolumn{2}{|c|}{\textbf{Total}} & 
      5.09 & 
      792.12 \\
      \hline
    \end{tabular}%
  }
\end{table}

\begin{figure}[t]
  \centering
  \includegraphics[width=\linewidth]{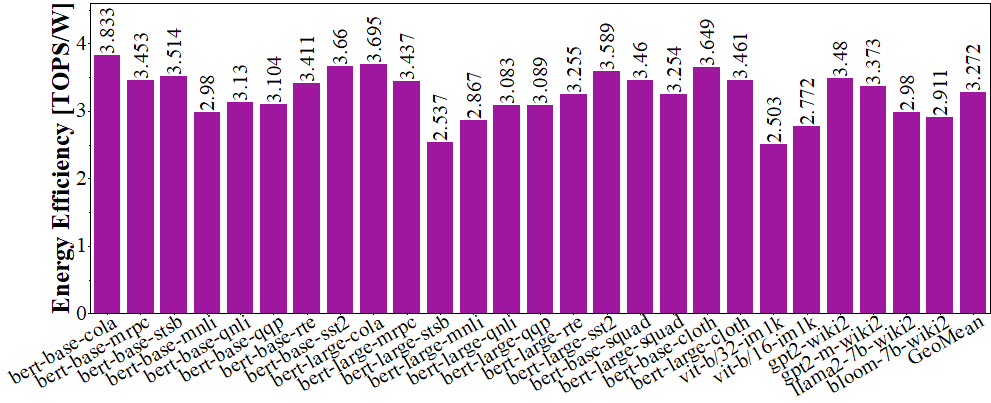}
  \caption{End-to-End energy efficiency of ESACT.}
  \label{fig:energy}
\end{figure}

\textbf{Area and Energy Analysis.} 
Table~\ref{tab:area} summarizes the breakdown of ESACT's area and power consumption. 
The total area and power of ESACT are 5.09\,mm\textsuperscript{2} and 792.12\,mW, respectively. 
The sparsity prediction module accounts for only 4.52\% of the total area and 7.25\% of the total power, 
demonstrating that SPLS can introduce substantial sparsity with relatively low hardware cost.
This efficiency stems from a hardware-aware design that 
exploits the properties of HLog quantization to construct a bit-level prediction unit. 
By leveraging the correlation among adjacent bits, 
ESACT enables efficient quantization while avoiding complex comparison operations. 
Moreover, attention prediction is performed using simple additions instead of costly multiplications, 
substantially reducing area and power consumption.

The dynamic allocation strategy utilizes counters to assign critical paths and controls data transmission in the FIFO based on similarity. 
Since the corresponding computations are still performed within the PEs, 
the control logic required for dynamic allocation is integrated into the PE Array's ctrl. 
The additional addition operations can be reused from part of the Softmax units within the Functional Module. 
As components such as FIFOs contribute negligibly to the overall power and area, they are collectively represented as ''others" within the Functional Module.
Figure~\ref{fig:energy} reports the end-to-end energy efficiency of ESACT across various datasets. 
On average, ESACT achieves an end-to-end energy efficiency of 3.27\,TOPS/W through 
joint optimization of QKV, attention, and FFN computations.

\begin{table}[t]
  \renewcommand{\arraystretch}{1.5} 
  \centering
  \caption{Area and Power Comparison of Quantization Methods Used in Different Accelerators}
  \label{tab:quant_area}
  \resizebox{\columnwidth}{!}{%
    \begin{tabular}{|c|c|c|c|}
      \hline
      \textbf{Quantization Method} & \textbf{Parameter} & \textbf{Area (mm\textsuperscript{2})} & \textbf{Power (mW)} \\
      \hline
      Sanger~\cite{sanger} & 
      \parbox{3cm}{\vspace{0.5em}\centering 8$\times$128 4-bit Multipliers \\ Adder Tree\vspace{0.5em}} & 
      0.23 & 
      81.70 \\
      \hline
      FACT~\cite{fact} & 
      \parbox{3cm}{\vspace{0.5em}\centering 128 LDZ Detectors \\ 8$\times$128 Adders \\ One-Hot Adder\vspace{0.5em}} & 
      0.14 & 
      37.98 \\
      \hline
      Enhance~\cite{enhance} & 
      \parbox{3cm}{\vspace{0.5em}\centering 128 Position Detectors \\ 8$\times$128 Adders \\ Adder Tree\vspace{0.5em}} & 
      0.26 & 
      80.76 \\
      \hline
      ESACT & 
      \parbox{3cm}{\vspace{0.5em}\centering 128 Shift Detectors \\ 8$\times$128 Adders \\ Converter\vspace{0.5em}} & 
      0.17 & 
      48.21 \\
      \hline
    \end{tabular}%
  }
\end{table}

\textbf{Area and Power Comparison of Quantization Methods Used in Different Accelerators.}
Table~\ref{tab:quant_area} summarizes the area and power consumption of various quantization methods 
adopted in recent accelerator designs under 28nm.
Compared to Sanger's 4-bit quantization, our method reduces area by 26\% and power by 41\%, 
indicating significantly lower overhead during the sparsity prediction stage.
Although our approach introduces a 21\% area increase and a 27\% power increase over 
FACT's PoT-based design that uses LDZ detectors to identify the first leading one, 
we achieve higher sparsity under higher accuracy constraints.
Enhance employs APoT quantization, which encodes each input as the sum of two one-hot vectors 
by detecting the positions of the top three leading ones.
We implement this scheme during the prediction stage using position detectors.
However, despite replacing multipliers with additions, APoT does not lead to power savings compared to 4-bit quantization.
This is because the transformation process itself still consumes over 40\% 
of the original multiplication energy~\cite{isscc,shiftaddllm}.
Moreover, unlike PoT and HLog quantization, the resulting quantized patterns in APoT lack structural regularity, 
making them ill-suited for efficient accumulation.
As a consequence, accumulation must be performed via adder trees, which introduces further area overhead.

\subsection{Comparison with Other Attention Accelerators}

To enable a fair comparison with existing attention accelerators, 
we scale the power, area, and throughput of SpAtten and Sanger to a 28nm technology node 
using the methodology outlined in~\cite{scale}. 
Table~\ref{tab:accelerators} summarizes the resulting energy and area efficiencies.
ESACT achieves an energy efficiency of 6677\,GOPS/W, 
surpassing SpAtten and Sanger by 2.95$\times$ and 2.26$\times$, respectively. 
This improvement stems from our SPLS mechanism, 
which enables both intra- and inter-row attention sparsity with lower power overhead.
Notably, SPLS enables over 50\% inter-row sparsity, 
significantly reducing redundant computation and contributing to the improved energy efficiency.
In terms of area efficiency, ESACT reaches 1039\,GOPS/mm\textsuperscript{2}, 
matching that of Sanger and exceeding SpAtten by 1.53$\times$. 
These results highlight the effectiveness of SPLS in achieving both energy-efficient and area-efficient attention acceleration.

\begin{table}[t]
  \renewcommand{\arraystretch}{1.5} 
  \centering
  \caption{Comparison of ESACT and Other Attention Accelerators}
  \label{tab:accelerators}
  \resizebox{\columnwidth}{!}{%
    \begin{tabular}{|c|c|c|c|}
      \hline
      \textbf{Accelerator} & \textbf{SpAtten~\cite{spatten}} & \textbf{Sanger~\cite{sanger}} & \textbf{ESACT} \\
      \hline
      \textbf{Sparse Attention Accuracy Loss} &
      0.7\% & 0.1\% & 0.2\%\\
      \hline
      \textbf{Tech (nm)} &
      40 & 55 & 28\\
      \hline
      \textbf{Freq (Hz)} &
      1G & 500M & 500M\\
      \hline
      \textbf{Area (mm\textsuperscript{2})} &
      1.55 & 16.9 & 5.09 \\
      \hline
      \textbf{Power (W)} &
      0.325 & 2.76 & 0.792 \\
      \hline
      \textbf{Attention Throughput (GOPS)} &
      360 & 2116 & 5288 \\
      \hline
      \textbf{Norm. Energy Effi. (GOPS/W)} &
      2261 & 2958 & 6677 \\
      \hline
      \textbf{Norm. Area Effi. (GOPS/mm\textsuperscript{2})} &
      677 & 1025 & 1039 \\
      \hline
    \end{tabular}%
  }
\end{table}

\section{CONCLUSION}\label{sec:conclusion}

We propose ESACT, a dedicated accelerator that enables end-to-end sparse acceleration of Transformers through the SPLS mechanism, 
which performs attention prediction and similarity detection. 
To support efficient hardware deployment, 
we introduce a bit-level prediction unit for lightweight sparsity estimation, 
along with a progressive generation scheme and dynamic allocation strategy to boost overall throughput. 
Compared to state-of-the-art attention accelerators SpAtten and Sanger, 
ESACT improves attention energy efficiency by 2.95$\times$ and 2.26$\times$, respectively.

\section*{Acknowledgments}
The authors would like to acknowledge the support from the Big Data Computing Center of Southeast University.


\bibliographystyle{IEEEtran}
\bibliography{refs}

\end{document}